%% file: preprint.tex
\documentclass[11pt, letterpaper, logo]{appier-ai-research}

\PassOptionsToPackage{comma,numbers,sort,compress}{natbib}

\usepackage[comma,numbers,sort,compress]{natbib}

\usepackage{hyperref}[citecolor=magenta]
\usepackage{fdsymbol}
\usepackage{enumitem}
\usepackage{makecell}
\hypersetup{
    colorlinks = true,
    citecolor = {magenta},
}

\usepackage{lipsum}

\pdftrailerid{redacted}

\makeatletter
\renewcommand\bibentry[1]{\nocite{#1}{\frenchspacing\@nameuse{BR@r@#1\@extra@b@citeb}}}
\makeatother

\usepackage{kantlipsum, lipsum}
\usepackage{dsfont}
\usepackage{gdm-colors}
\usepackage{subfigure}
\usepackage{bbm}
\usepackage{wrapfig}
\usepackage{multirow}
\usepackage{graphicx}
\usepackage{tabularx}
\usepackage{fancyvrb}
\usepackage{longtable}

\usepackage[most,skins,theorems]{tcolorbox}

\tcbset{
  aibox/.style={
    width=\linewidth,
    top=8pt,
    bottom=4pt,
    colback=blue!6!white,
    colframe=black,
    colbacktitle=black,
    enhanced,
    center,
    attach boxed title to top left={yshift=-0.1in,xshift=0.15in},
    boxed title style={boxrule=0pt,colframe=white,},
  }
}
\newtcolorbox{AIbox}[2][]{aibox,title=#2,#1}
\newenvironment{observation}{\par\smallskip\noindent\textbf{Observation.}\itshape}{\par\smallskip}
\newenvironment{pitfall}{\par\smallskip\noindent\textbf{Pitfall.}\itshape}{\par\smallskip}
\newenvironment{lesson}{\par\smallskip\noindent\textbf{Lesson.}\itshape}{\par\smallskip}
\definecolor{lightblue}{rgb}{0.22,0.45,0.70}

\usepackage{algorithm}
\usepackage{algpseudocode}
\usepackage{seqsplit}
\tcbset{
  aiboxc/.style={
    colback=blue!6!white,
    colframe=black,
    colbacktitle=black,
    enhanced jigsaw,
    breakable,
    center,
    attach boxed title to top left={yshift=-0.1in,xshift=0.15in},
    boxed title style={boxrule=0pt,colframe=white,},
  }
}
\newenvironment{AIboxC}[2][]{%
  \def\AIboxCStoredTitle{#2}%
  \begin{tcolorbox}[aiboxc,title=\AIboxCStoredTitle,#1]%
}{%
  \end{tcolorbox}%
}
\DeclareFloatingEnvironment{boxes} 
\newcommand{\boxref}[1]{\hyperref[{#1}]{TextBox~\ref*{#1}}}
 
\newcommand{\eg}{\textit{e}.\textit{g}.,\ }

\graphicspath{{figures/}}

\title{Joint Optimization of Tool Creation and Use for Large Language Model Agents}

\correspondingauthor{ray.tam@appier.com}

\author[1,2]{Zhi Rui Tam}
\author[1]{Chieh-Yen Lin}
\author[2]{Yun-Nung Chen}
\author[1,2]{Shao-Hua Sun}
\author[2]{Hung-yi Lee}

\affil[1]{Appier AI Research}
\affil[2]{National Taiwan University}

\begin{abstract}
Tool-augmented language models are bounded by the APIs humans bothered to write; existing tool-creation systems patch this by prompting a frozen LLM at inference time, leaving the model that writes a tool decoupled from the one that uses it, with no signal that the schemas it produces are schemas it can invoke.
We propose \textbf{SMITH} (Schema-grounded Multi-task Iterative Tool Honing), a reinforcement learning framework that jointly trains tool creation and tool use inside a single policy. Each rollout is either a \emph{build} task (write a tool from a few examples) or a \emph{use} task (invoke a pooled tool on a held-out question). Three separate reward axes catch schema, code, and outcome failures independently, so each failure mode contributes its own gradient.
A 4B Qwen3 trained with SMITH on 13 procedural reasoning tasks with exact verifiers reaches $79.8$ macro-average accuracy on held-out tasks, the best across all evaluated methods and ahead of an untrained 30B-A3B tool-writer. It also reaches $40.4$ on TabMWP-Hard and $42.6$ on out-of-domain GQA ($+7.6$ over the best same-backbone inference-time baseline), without any visual or tabular training data. Tools written by our 4B models also lifted the performance of LFM-2.5-350M and Qwen3-30B-A3B under same reasoning tasks.
\par
\centering
  \vspace{0.8em}
  \includegraphics[height=1.2em]{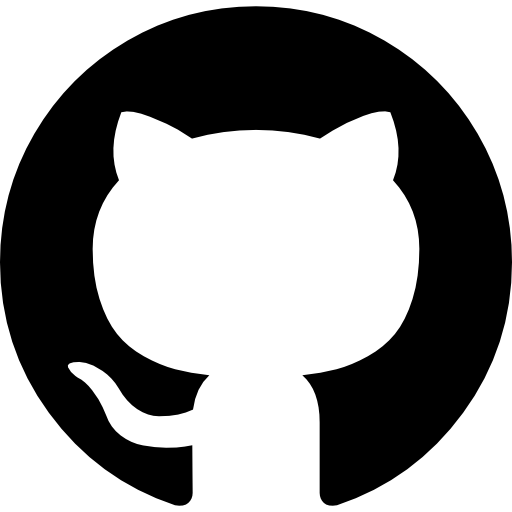}~
  \href{https://github.com/appier-research/smith}{\texttt{Code}}~~
  \includegraphics[height=1.2em]{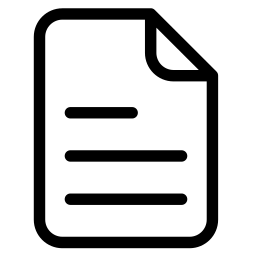}~
  \href{https://tool-use-smith.github.io/}{\texttt{Project Page}}
  \vspace{-0.8em}
\par
\end{abstract}

\begin{document}

\maketitle

\section{Introduction}
\input{sections/intro_sun}

\section{SMITH: Schema-grounded multi-task iterative tool honing}
\input{sections/smith}

\section{Experiments}
\input{sections/experiment_setup}
\input{sections/results}

\section{Limitations and discussions}
\label{sec:limitations}
\input{sections/limitation}

\section{Conclusion}
\input{sections/conclusion}

\input{sections/ack}

\renewcommand{\bibname}{References}
\bibliographystyle{plainnat}
\bibliography{references}

\newpage 

\appendix 
\part*{Appendices}

\section{Related works}
\label{sec:related_works}
\input{sections/related_work}
\input{appendix/verify_and_llm_judge}

\input{appendix/judge_prompt}
\input{appendix/prompt_templates}
\input{appendix/tool_pool_design}
\input{appendix/difficulty_split_table}
\input{appendix/ood_inference_protocol}
\input{appendix/gqa_visual_tool_setup}
\input{appendix/retool_evaluation}
\input{appendix/latm_distillation}
\input{appendix/baseline_agentic_comparison}
\input{appendix/baseline_failure_analysis}
\input{appendix/tabmwp_hard_dataset}
\input{appendix/compute_cost}
\input{appendix/trove}

\end{document}

%% file: sections/intro_sun.tex
Human progress depends on accumulated tools. Rather than solving every problem from first principles, people rely on instruments refined over generations~\citep{hutchins1995cognition, clark1998extended}. Large Language Models (LLMs) face a similar limitation: relying only on parametric memory restricts their ability to perform exact computation~\citep{cobbe2021training}, access up-to-date knowledge~\citep{chengdated}, and carry out reliable symbolic reasoning~\citep{goutora}. Tool-augmented LLMs were introduced to overcome these limits~\citep{komeili2022internet, thoppilan2022lamda, taylor2022galactica}. By invoking external interfaces such as calculators, code interpreters, and search engines, models can solve problems beyond what is encoded in their frozen weights.


However, existing tool-augmented systems still rely on fixed, human-designed toolsets. Such predefined APIs may be incomplete, poorly matched to new tasks, or entirely unavailable, placing a hard limit on model capability. This motivates a shift from static tool use to \emph{dynamic tool creation}, where models synthesize reusable callable functions on demand. LATM~\citep{cai2023large} first explored this direction by prompting GPT-4 to generate JSON-schema tools from demonstrations, and later work incorporated retrieval, verification, and multi-stage generation pipelines~\citep{yuancraft, wang2024trove, ma2025automated}. Despite this progress, existing methods do not explicitly optimize for tool quality or reusability during training of LLMs. Moreover, they separate tool creation from tool use: a stronger model writes the tool while a weaker model invokes it~\citep{cai2023large, yuancraft}. As a result, the tool creator is never incentivized to design schemas it can reliably use itself. A key open problem is therefore \textit{how to train a single model to become both a better tool creator and a better tool user}. 

Reinforcement learning provides a natural framework for this objective~\citep{bai2022training, shao2024deepseekmath, yu2025dapo}: a model can create a tool, apply it to held-out queries, and directly optimize against answer correctness. Yet, training under this paradigm introduces two major challenges. First, \textbf{reward decomposition}: a generated tool contains both a \emph{schema} (\eg function name, parameters, and types) and a \emph{backend implementation} (\eg executable Python code). Failures in these two components require different corrective signals, making reward assignment nontrivial. Second, \textbf{circular evaluation}: evaluating tool quality requires a judge, but self-evaluation is inherently unreliable~\citep{panickssery2024llm}, while fixed external evaluators raise questions about trustworthiness and alignment~\citep{zheng2023judging}. In addition, a competent evaluator must itself understand tool use well enough to assess whether a schema is practically callable and reusable across harder downstream queries.

To address these challenges, we propose \textbf{SMITH} (\textbf{Schema-grounded} \textbf{Multi-task} \textbf{Iterative} \textbf{Tool} \textbf{Honing}), a reinforcement learning framework that jointly trains tool creation and tool use within a single policy. SMITH disentangles schema and implementation failures through three complementary reward signals: execution accuracy, LLM-as-judge quality, and format consistency. To reduce circular evaluation, the judge is periodically synchronized from the evolving policy rather than sharing live training weights, enabling the evaluator to improve alongside the policy while maintaining stability. Finally, SMITH evaluates tools created on simpler tasks using harder downstream queries, explicitly rewarding reusable abstractions rather than task-specific shortcuts.

We train SMITH on 13 procedural task families from Reasoning-Gym~\citep{stojanovski2025reasoning} and evaluate both in-domain generalization and zero-shot transfer to unseen benchmarks. Our 4B model achieves the best overall held-out Reasoning-Gym performance ($79.8\%$ macro accuracy), outperforming inference-time tool-writing frameworks such as LATM~\citep{cai2023large}, CRAFT~\citep{yuancraft}, Trove~\citep{wang2024trove}, and KTCE~\citep{ma2025automated}, as well as models distilled from substantially larger teachers. Despite using far fewer decoding tokens, SMITH also surpasses a 30B inference-time tool writer on unseen tasks. Beyond procedural reasoning, SMITH transfers zero-shot to entirely new modalities: it achieves state-of-the-art performance on TabMWP-Hard and improves GQA visual question answering by up to $+7.6$ points over same-backbone baselines, despite never being trained on tabular or visual data. The learned tools further generalize across models: tools written by our 4B policy enable a frozen 350M model to match the performance of a 30B tool writer. Finally, the same training recipe consistently improves both Qwen3-8B and Granite-3.3-8B, suggesting that jointly optimizing tool creation and tool use under structured rewards is a scalable path toward reusable and transferable tool-building capabilities.

%% file: sections/smith.tex
\begin{figure}[t!]
\includegraphics[width=\linewidth]{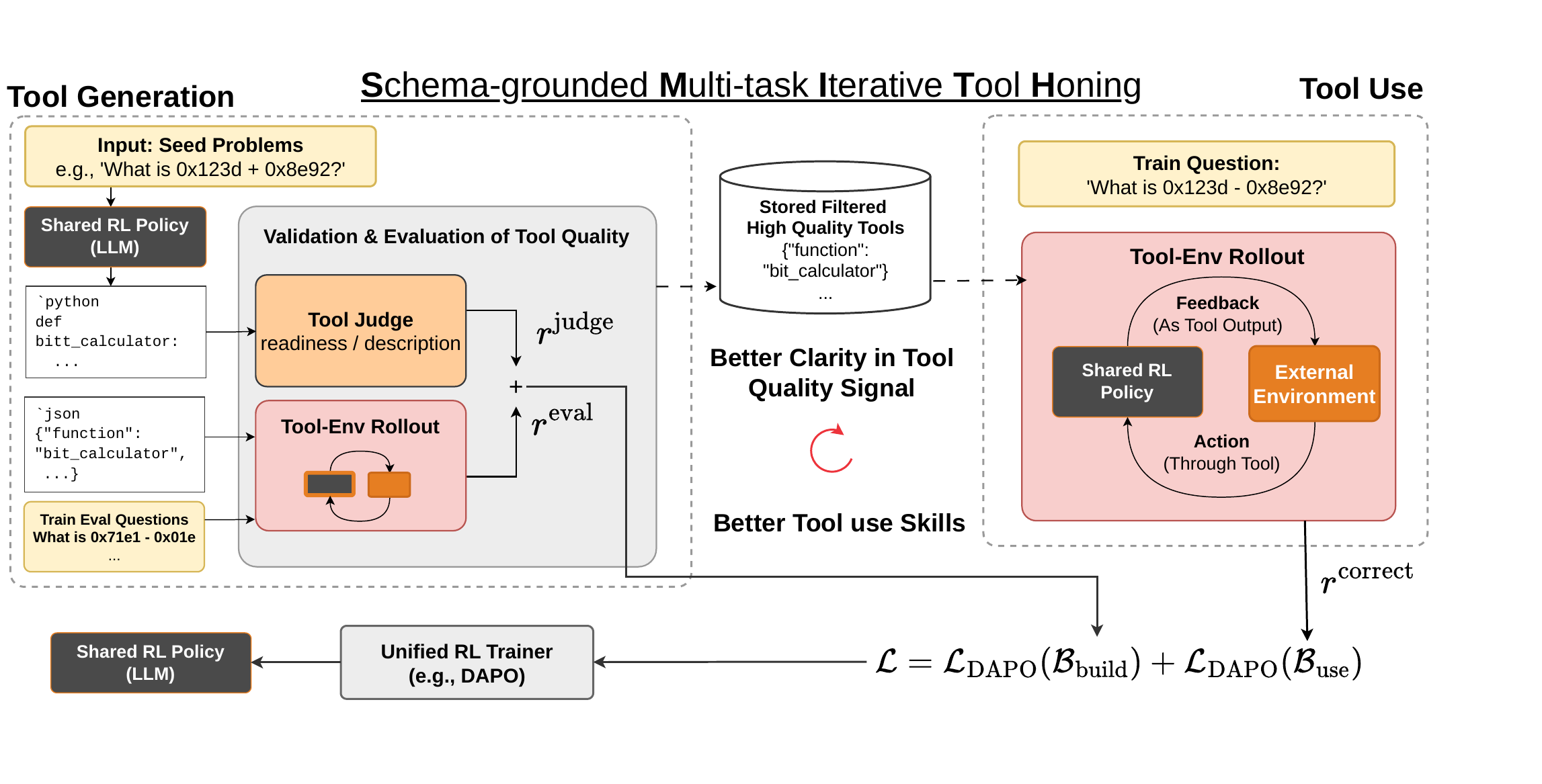}
\vspace{-8mm}
\caption{SMITH jointly trains tool creation and tool use in a single shared policy. The build task rewards schema correctness and execution accuracy; the use task rewards answer correctness using only the schema. Because the same model that writes a tool must also invoke it, ambiguous or broken schemas are penalized directly---a feedback loop that prompting-only approaches cannot provide.}
\label{fig:joint_training}
\vspace{-2mm}
\end{figure}

SMITH is a \emph{multi-task} RL framework that jointly trains two complementary skills: \emph{tool creation} (build) and \emph{tool use}.
In the build task the model synthesizes a reusable tool, expressed as both a Python function and an OpenAI-compatible JSON schema, a structured description of the function's name, parameters, and types that exposes the tool through a standard invocation interface.
In the use task the model sees \emph{only} this compact JSON schema (not the underlying code) and must invoke the tool to answer a held-out question.
This schema-grounded design forces the model to produce concise, self-contained interfaces: a tool whose schema is ambiguous or incomplete will fail at use time, providing a direct training signal for schema quality.

We train the shared policy using DAPO~\citep{yu2025dapo}, a clip-higher variant of GRPO~\citep{shao2024deepseekmath} that stabilizes entropy and avoids reward
collapse during on-policy rollouts.
Each training step samples a batch $\mathcal{B}$ of prompts split into two equal halves, $|\mathcal{B}_{\mathrm{build}}| = |\mathcal{B}_{\mathrm{use}}| = B/2$.
Keeping the two reward streams structurally separate while training on the same policy allows the two skills to reinforce each other.
\paragraph{Task formulation.} For each training instance of tool generation, the policy receives $N{=}4$ question-answer pairs $\{(q_i, a_i)\}_{i=1}^{N}$ as context for \emph{problem induction}: the model must infer a general solution strategy and express it as both a Python function $\mathcal{C}$ and an OpenAI-compatible JSON schema $\mathcal{S}$.
A disjoint set of $K{=}16$ held-out questions $\mathcal{T} = \{(q_j, a_j)\}_{j=1}^{K}$ is then used to evaluate the generated tool without the model ever observing the ground truth answers at generation time.
For use tasks, the model instead receives a single target question and must invoke a tool from the pool (if one exists for the category) or first build one from the same $N$ in-context examples before invoking it.

\subsection{Build task rewards}
\label{sec:build}

\paragraph{Evaluation reward.}
The generated tool $(\mathcal{C}, \mathcal{S})$ is evaluated against the hidden test set $\mathcal{T}$.
Each question $q_j$ is presented to an evaluator model $\pi^{\mathrm{eval}}$ that may call the generated tool; a correct answer requires both a successful tool invocation and an output verified by an LLM equivalence judge.
$\pi^{\mathrm{eval}}$ is initialized from the same base checkpoint as the policy and is periodically refreshed by copying the latest policy weights, providing a stable but improving evaluation target without the instability of evaluating against the live training weights.
The evaluation reward is the fraction of test questions answered correctly:
\begin{equation}
  r^{\mathrm{eval}}
  = \frac{1}{|\mathcal{T}|} \sum_{j=1}^{|\mathcal{T}|}
    \mathbf{1}\!\left[
      \pi^{\mathrm{eval}}\!\left(q_j \mid \mathcal{C}, \mathcal{S}\right) \approx a_j
    \right].
\end{equation}
Counting only answers obtained through a successful tool call prevents the policy from exploiting the fallback of text-only reasoning, which would bypass the tool-writing objective entirely.

\paragraph{Format reward.}
We apply a format reward $r^{\mathrm{fmt}} \in \{0, r_{f}\}$ ($r_f = 0.5$) when the response contains exactly one Python block and one JSON block whose function names and parameter signatures are mutually consistent.
If the generated response cannot be parsed into a valid $(\mathcal{C}, \mathcal{S})$ pair (e.g.\ missing code or schema), the rollout is terminated early and all reward axes are set to zero; $r^{\mathrm{fmt}} = 0$ alone does not terminate the rollout.
The build environment reward folds the format and evaluation signals together:
\begin{equation}
  r^{\mathrm{env}}_{\mathrm{build}} = r^{\mathrm{fmt}} + r^{\mathrm{eval}}.
  \label{eq:build_env}
\end{equation}

\paragraph{Judge reward.}
A separate LLM judge $\pi^{\mathrm{judge}}$ scores the generated tool on three
axes: code correctness $s_{\mathrm{code}}$, schema quality $s_{\mathrm{schema}}$,
and an overall quality score $s_{\mathrm{overall}} \in [0, 1]$.
A schema-code alignment check is applied after scoring: if the function signatures in $\mathcal{C}$
and $\mathcal{S}$ disagree, the score is halved; a syntax error in $\mathcal{C}$ yields a fixed
negative reward to penalize broken code:
\begin{equation}
  r^{\mathrm{judge}} =
  \begin{cases}
    -0.5                         & \text{if } \mathcal{C} \text{ contains a syntax error,} \\
    0.5 \cdot s_{\mathrm{overall}} & \text{if schema and code signatures disagree,} \\
    s_{\mathrm{overall}}           & \text{otherwise.}
  \end{cases}
\end{equation}
Crucially, $r^{\mathrm{judge}}$ is \emph{not} folded into $r^{\mathrm{env}}_{\mathrm{build}}$;
it is passed to DAPO as an independent reward axis with its own coefficient, keeping the
execution signal and the semantic quality signal disentangled.
The build task therefore contributes two independent reward axes to DAPO:
\begin{equation}
  \bigl(\,r^{\mathrm{env}}_{\mathrm{build}} = r^{\mathrm{fmt}} + r^{\mathrm{eval}},\quad r^{\mathrm{judge}}\,\bigr),
  \label{eq:build_axes}
\end{equation}
exposing format consistency, execution accuracy, and judge quality as the three reward signals that govern build-task learning, complemented by a single correctness axis from the use task.
Any tool with $r^{\mathrm{eval}} > 0$ is added to the shared Tool Pool $\mathcal{P}$ for reuse
in subsequent use-task rollouts (Section~\ref{sec:use}).
\subsection{Use task rewards}
\label{sec:use}
For use tasks the model executes a multi-turn dialogue with the tool for up to $T{=}5$ turns,
terminating early when the model produces a final answer or fails to emit a valid tool call.
\paragraph{Correctness reward.}
Let $c \in \{0,1\}$ indicate whether the final answer matches the ground truth $a^{*}$, verified by string normalization, with an LLM equivalence judge as a fallback.
To penalize turn-budget exhaustion, the correctness score is scaled by an efficiency multiplier $\eta(\rho)$ that decays as the turn fraction $\rho = \min(n/T, 1)$ grows.
We define a minimum reward floor $\eta_{\min}$ to ensure the policy is never strictly indifferent to correctness, even at the turn limit:
\begin{equation}
  r^{\mathrm{correct}} = 2\,c \cdot \eta(\rho), \qquad
  \eta(\rho) =
  \begin{cases}
    1 - 2(1 - \eta_{\text{mid}})\rho & \rho \le 0.5, \\
    \max\!\bigl(\eta_{\min},\; \eta_{\text{mid}}\,(1 - 2(\rho-0.5))^2\bigr) & \rho > 0.5.
  \end{cases}
  \label{eq:correct}
\end{equation}
The piecewise form is continuous at $\rho = 0.5$ by construction: both branches evaluate to $\eta_{\text{mid}}$ at the boundary. We set $\eta_{\min} = 0.3$ and $\eta_{\text{mid}} = 0.7$; with these values the floor activates at $\rho \approx 0.83$ (after 5 turns). These values were chosen based on small-scale preliminary runs and held fixed throughout all experiments.
The use task therefore contributes a single reward axis to DAPO : $r^{\mathrm{correct}}$
the efficiency-weighted final-answer correctness defined in Eq.~\ref{eq:correct}; tool-execution success is implicit, since correctness requires a valid tool invocation.
Together with the two build axes (Eq.~\ref{eq:build_axes}), this gives three independent reward axes for joint policy optimization.
\paragraph{Joint policy update.}
Each training batch is a disjoint union $\mathcal{B} = \mathcal{B}_{\mathrm{build}} \sqcup \mathcal{B}_{\mathrm{use}}$ with $|\mathcal{B}_{\mathrm{build}}| = |\mathcal{B}_{\mathrm{use}}| = B/2$, where $B$ is the per-iteration batch size. This keeps the gradient contribution from each task balanced.
DAPO computes per-prompt advantages independently within each generation group and accumulates gradients across both task types in a single backward pass:
\begin{equation}
  \mathcal{L} = \mathcal{L}_{\mathrm{DAPO}}(\mathcal{B}_{\mathrm{build}})
              + \mathcal{L}_{\mathrm{DAPO}}(\mathcal{B}_{\mathrm{use}}).
  \label{eq:joint_loss}
\end{equation}
Because both losses share the same policy parameters $\theta$, gradients from build and use prompts jointly update the model in every step, with no explicit advantage combination across task types.

%% file: sections/experiment_setup.tex
\subsection{Training tasks}
\label{sec:training_tasks}

We train on \textbf{13 task categories} from Reasoning-Gym (RG)~\citep{stojanovski2025reasoning},
selected for three properties: answers are exact and automatically verifiable (enabling reward
computation without human annotation), each task exposes a curriculum of difficulty levels
(enabling the easy-to-hard protocol described in Sec.~\ref{sec:easy_hard}), and questions can be generated procedurally.
The categories span arithmetic (\textit{bitwise arithmetic}, \textit{cryptarithmetic}), algorithms (\textit{bit counting}, \textit{LCM}, \textit{GCD}, \textit{base conversion}, \textit{isomorphic string}), algebra (\textit{polynomial equations}, \textit{polynomial multiplication}), games (\textit{countdown}, \textit{Tower of Hanoi}), and logical reasoning (\textit{knights-and-knaves}, \textit{Caesar cipher}).
This design separates tool-writing quality from instance difficulty: a correct tool induced on easy
examples must generalize to hard examples without further adaptation, a separation unavailable in
fixed-label benchmarks where harder instances cannot be generated on demand.

\subsection{Evaluation benchmarks}
\label{sec:benchmarks}

We measure transfer at three levels of increasing distance from training : RG (Seen) and RG (Unseen).

\textbf{RG (Seen).} Macro-average accuracy over the 13 training task categories, evaluated at the hardest curriculum level. This measures whether the learned policy produces tools that generalize beyond easy induction contexts to harder instances of the same task families.

\textbf{RG (Unseen).} Macro-average accuracy over \textbf{10 RG tasks} withheld entirely from RL training, spanning: arithmetic (\textit{calendar arithmetic}, \textit{complex
arithmetic}, \textit{time intervals}); algebra (\textit{Chinese remainder theorem}, \textit{simple equations}); algorithms (\textit{group anagrams}); and logic \& games (\textit{ab}, \textit{self-referential sequence}, \textit{syllogism}, \textit{Puzzle-24}). 
While these categories appear in RG (Seen), the specific problem types are entirely absent from training, probing cross-task generalization within Reasoning-Gym.

\textbf{Out-of-domain benchmarks.} \textbf{TabMWP-Hard}: a strengthened variant of TabMWP~\citep{lu2022dynamic}; the original is trivially solved in CoT (96.8\% EM) because tables average $<\!10$ rows and 2 columns, so we extend rows to up to 5{,}000 and add unrelated columns (Appendix~\ref{app:tabmwp_hard}). \textbf{GQA}~\citep{hudson2019gqa}: visual question answering, requiring visual tools. Together these cover the three most relevant tool-use classes (computation, knowledge access, non-textual modality) per \citet{wang2024tools}.

\subsection{Models and training configuration}
\label{sec:impl}

We use \textsc{Qwen3-4B-Instruct}~\citep{yang2025qwen3} as the primary target for its
strong instruction-following and tool use ability. For all training, we fine-tune with LoRA
($r{=}64$, $\alpha{=}128$) to reduce the training compute.
We train with DAPO~\citep{yu2025dapo} for 60 gradient steps across 13 task categories, with generation and use tasks at a 1:1 ratio within each batch, $n_{\mathrm{gen}}{=}8$ rollouts per prompt, temperature $0.7$, $\beta{=}0.01$ (KL coefficient), learning rates $6{\times}10^{-5}$ (4B) and $1{\times}10^{-5}$ (8B). 
The Tool Pool (Appendix~\ref{app:tool_pool}) caches up to 20 verified tools per category; each
use-task rollout injects 1 domain tool (from the correct category) and 2 distractor tools
(from unrelated categories, forcing the model to identify and invoke the correct schema) into
the prompt.

\subsection{Train/test split design for easy-to-hard transfer}
\label{sec:easy_hard}

We separate the difficulty of tool induction from the difficulty of tool evaluation.
For each RG task, the \emph{induction context} is drawn from the easier curriculum bands (train), while the \emph{tool evaluation set} is drawn from the hardest band (test); the exact mapping is task-specific.
The reward $r^{\mathrm{eval}}$ (Sec.~\ref{sec:build}) is therefore sharply reduced for any tool that merely pattern-matches the easy induction context without abstracting the underlying algorithm, pushing the policy toward tools that are concise, interpretable, and reusable.
Unlike prior inference-time tool-generation frameworks~\citep{cai2023large, yuancraft, wang2024trove, ma2025automated}, which evaluate tools on the same difficulty distribution used to create them, our protocol forces an easy-to-hard generalization gap at every step.

The complete per-task split for all 13 categories is in Appendix~\ref{app:difficulty_split}.

\subsection{Baselines}
\label{sec:baselines}

All baselines use \textsc{Qwen3-4B-Instruct} unless noted. \textbf{Standard CoT} provides the no-tool ceiling. However, in GQA since LLMs cannot process visual information without external visual tool, the LLMs can only answer based on its knowledge learned from textual world.

\textbf{LATM}~\citep{cai2023large}, \textbf{CRAFT}~\citep{yuancraft},
\textbf{Trove}~\citep{wang2024trove}, and \textbf{KTCE}~\citep{ma2025automated} all create tools
at inference time with a frozen Qwen3-4B-Instruct backbone, differing only in scaffolding
complexity; comparing them on the same backbone tests whether RL training adds value beyond
prompt engineering.\footnote{We rewrite the LATM prompt for better schema clarity.}
We additionally evaluate \textbf{LATM with Qwen3-30B-A3B} as a deliberate scaling probe, isolating whether sheer model size on the same inference-time harness can close the gap to RL training.
\textbf{ReTool~(4B distill Qwen-32B)}~\citep{feng2025retool} is a multi-turn code-execution policy (up to 10 turns) distilled from Qwen-32B traces that never produces a reusable schema; including it isolates the contribution of the schema-grounded tool representation, since both approaches use execution feedback but only SMITH produces reusable callable schemas.
\textbf{LATM~(4B distill GPT-4.1)} fine-tunes the backbone on tool-writing trajectories generated by GPT-4.1; comparing against it tests whether RL training over an explicit reward signal yields better tools than behavioral cloning from a stronger frozen oracle.

\paragraph{Evaluation protocol.}
We report macro-average accuracy with two Reasoning-Gym averages: \emph{RG (Seen)} over the 13 training categories and \emph{RG (Unseen)} over the 10 held-out categories. 

%% file: sections/results.tex
\subsection{Main results}

We compare SMITH against inference-time tool-creation baselines (LATM, CRAFT, TroVE, KTCE),
distillation baselines (ReTool 4B distilled from Qwen-32B; LATM 4B distilled from GPT-4.1),
and a larger model (Qwen3-30B-A3B) on LATM, all on the same evaluation protocol.
Two findings stand out in Table~\ref{tab:rg_results}.
First, against distillation: SMITH (4B, RL) generalizes more reliably than 4B models
distilled from much larger oracles, attaining the highest held-out RG accuracy overall
($79.9$) versus $63.2$ for ReTool and $65.8$ for LATM-distilled. ReTool in particular leads
on RG (Seen) ($92.2$ vs.\ our $85.2$) but loses nearly $30$ points on RG (Unseen),
indicating that rejection-sampled distillation overfits to the demonstrator's training
distribution rather than learning a transferable build-and-use policy.
Second, against more elaborate scaffolds: SMITH's simple scaffold LATM is simply write multiple versions of tools and then pick the best, is the strongest results on RG (Seen) at
$85.2$ and beats every scaffolding baseline on RG (Unseen), including CRAFT ($76.5$), KTCE ($65.1$), and TroVE ($55.9$). In addition to beating larger models Qwen3-30B-A3B Instruct and a distilled version of Qwen3 4B from GPT-4.1 responses through rejection sampling finetuning.
These results suggest that learning \emph{how} to build a tool from a verifiable reward outperforms both behavioral cloning from a stronger oracle and hand-engineered retrieval/refinement loops.

Third, on token efficiency: SMITH achieves the strongest aggregate accuracy with the smallest output budget at $100$ tokens on average, roughly $32\times$ fewer than Standard CoT ($3{,}206$) and $6\times$ fewer than ReTool ($633$), while input usage remains modest at $664$ tokens, comparable to LATM ($607$) and well below CRAFT ($1{,}226$) and ReTool ($1{,}707$). The asymmetry is informative: scaffolding baselines such as CRAFT and the distillation-trained ReTool spend far more input tokens on retrieved exemplars or stitched prompts, and CoT spends its budget on long unconditioned reasoning, yet none recovers SMITH's holdout accuracy. This shows that the RL objective shifts the work out of decode-time reasoning and into reusable tool code, so each query is resolved by a short tool invocation rather than a long chain of thought.

\begin{table}[t]
  \centering
  \caption{Reasoning-Gym results (Qwen3-4B-Instruct); CRAFT, TroVE, KTCE, ReTool, LATM-distill, and SMITH report mean$\pm$std over seeded re-evaluations. SMITH leads held-out generalization (RG Unseen 79.9) over distilled models (ReTool 63.2, LATM-distill 65.8) and every scaffolding baseline, using 32$\times$ fewer output tokens than standard CoT. Fixing baseline grading bugs raises TroVE's RG (Unseen) from 40.6 to 55.9 and KTCE's from 48.7 to 65.1. \emph{RG (Seen/Unseen)}: macro-average over training/held-out tasks. $^*$LATM prompt rewritten for schema clarity. I/O: input\,/\,output tokens.}
  \label{tab:rg_results}
  \footnotesize
  \setlength{\tabcolsep}{3pt}
  \begin{tabular}{lcccccccc}
    \toprule
     & \multicolumn{1}{c}{\textbf{Seen}} & \multicolumn{6}{c}{\textbf{Unseen RG}} &  \\
    \cmidrule(lr){2-2}\cmidrule(lr){3-8}
    \textbf{Method} & \textbf{Avg} & \textbf{Logic} & \textbf{Game} & \textbf{Algebra} & \textbf{Arith} & \textbf{Algo} & \textbf{Avg} & \textbf{I/O} \\
    \midrule
    Standard CoT & 58.0 & 49.9 & 60.3 & 56.8 & 62.7 & 48.6 & 55.7 & 173 / 3,206 \\
    LATM*~\citep{cai2023large} & 77.6 & 53.9 & 55.5 & 38.6 & 53.0 & 90.2 & 58.3 & 607 / 174 \\
    LATM* - Qwen3-30B-A3B & 74.0 & \underline{68.7} & 64.2 & \underline{97.3} & 56.5 & 84.0 & 74.1 & 659 / 405 \\
    CRAFT~\citep{yuancraft} & 74.1$\pm$0.7 & 27.4$\pm$1.3 & \textbf{89.5}$\pm$\textbf{0.0} & 94.2$\pm$1.2 & \underline{76.6$\pm$1.1} & \underline{95.0$\pm$0.0} & \underline{76.5$\pm$0.4} & 1,226 / 418 \\
    Trove~\citep{wang2024trove} & 52.6$\pm$0.4 & 60.7$\pm$2.4 & 10.6$\pm$1.3 & 51.2$\pm$3.5 & 59.8$\pm$0.8 & \textbf{97.0}$\pm$\textbf{0.0} & 55.9$\pm$0.6 & 347 / 575 \\
    KTCE~\citep{ma2025automated} & 61.0$\pm$1.5 & 60.2$\pm$1.5 & \underline{79.8$\pm$1.6} & 70.6$\pm$0.3 & 45.6$\pm$0.4 & 69.3$\pm$1.8 & 65.1$\pm$0.2 & 319 / 404 \\
    \midrule
    \textbf{ReTool (distill Qwen-32B)} & \textbf{92.2}$\pm$\textbf{0.8} & 50.3$\pm$2.3 & 55.0$\pm$0.8 & 48.7$\pm$0.6 & \textbf{79.8}$\pm$\textbf{2.7} & 82.4$\pm$0.1 & 63.2$\pm$0.4 & 1,707 / 633 \\
    \textbf{LATM (distill GPT-4.1)} & 81.7$\pm$4.4 & 37.6$\pm$15.2 & 58.1$\pm$12.2 & 91.3$\pm$7.7 & 51.6$\pm$10.4 & 93.2$\pm$5.9 & 65.8$\pm$4.1 & 638 / 207 \\
    \midrule
    \textbf{SMITH} & \underline{85.2$\pm$2.7} & \textbf{74.2}$\pm$\textbf{0.6} & 63.7$\pm$1.1 & \textbf{97.9}$\pm$\textbf{2.6} & 70.6$\pm$2.1 & 93.0$\pm$0.4 & \textbf{79.9}$\pm$\textbf{2.2} & 664 / 100 \\
    \bottomrule
  \end{tabular}
\end{table}

For all baselines TroVE, CRAFT and KTCE we did manually inspect the original codebase and found some python code parsing issues when running on Qwen3-4B. Hence we have fixed those issues in our revision. We also tried to modify the TroVE original prompts and we found it did not perform any better or significantly worse than the original prompt.

\subsection{Tools transfer across model scale}

A natural question is whether the tools synthesized by SMITH encode genuinely generalized solutions or whether they only work with the same policy model that wrote them. We test both directions: pairing SMITH's 4B writer with a much smaller consumer and with a much larger one.

\textbf{Smaller student.} We pair our best fine-tuned 4B tool generation model with LFM2.5-350M~\citep{liquidai2025lfm2}, a 350\,M-parameter small model with strong tool use ability, to use those tools at inference time.
Table~\ref{tab:lfm25_scaling} shows that pairing LFM2.5-350M with our RL 4B tool-generation model increases holdout RG accuracy from $11.6$ to $42.9$, matching the much larger Qwen3-30B-A3B-Instruct writer ($41.5$); in training tasks, our 4B model ($21.1$) actually exceeds the 30B untrained model ($10.4$). We also train SMITH model using LFM2.5 as reward signal on tool generation, while it performs the best in RG Seen set but underperforms on RG Unseen.

\begin{table}[t]
  \centering
  \caption{Tools from SMITH's RL-trained 4B writer enable a 350M model (LFM2.5) to match a 30B tool writer on held-out tasks (42.9 vs.\ 41.5 RG Unseen), showing that SMITH's tools encode genuinely generalizable solutions that transfer across model families. First row is the no-tool baseline. RG (Seen)/(Unseen): macro-average accuracy (\%). Best per column in \textbf{bold}.}
  \label{tab:lfm25_scaling}
  \small
  \begin{tabular}{lcccc}
    \toprule
    \textbf{Method} & \textbf{RG (Seen)} & \textbf{RG (Unseen)} & \textbf{TabMWP} & \textbf{GQA} \\
    \midrule
    \textsc{LFM2.5-350M} (no tool) & 14.2 & 11.6 & 0.0 & 0.1 \\
    \midrule
    \quad\textsc{+} Qwen3-4B-Instruct (tool writer) & 36.8 & 23.4 & 4.2 & 0.1 \\
    \quad\textsc{+} Qwen3-30B-A3B (tool writer) & 10.4 & 41.5 & 4.1 & 0.1 \\
    \quad\textsc{+} RL on LFM2.5-350M  & \textbf{39.1} & 30.2 & 4.2 & 0.1 \\
    \quad\textsc{+} RL 4B (Ours) & \underline{38.9} & \textbf{42.9} & 4.1 & 0.1 \\
    \bottomrule
  \end{tabular}
\end{table}

\textbf{Larger consumer.} The complementary question is whether the 4B writer remains useful once a much stronger tool user is available, the practical deployment setting where one could simply let the larger model write its own tool instead. We pair tools generated by our RL-trained 4B writer with Qwen3-30B-A3B-Instruct as the tool \emph{consumer} and compare against LATM$^*$, where the same 30B model both writes and uses its own tool, across the 25 tasks the two configurations share (10 held-out RG, 13 seen RG, TabMWP-Hard, GQA).
Table~\ref{tab:30b_consumer} shows the SMITH-4B-written tools lift the 30B consumer on every group, most sharply on TabMWP-Hard ($0.7 \to 38.8$), and raise the task-weighted overall score from $70.2$ to $76.6$.
A stronger tool user therefore does not make its own self-written tool preferable: the RL-trained 4B writer's tools remain a better source of tools than the 30B model's own. Taken together, these two results show that SMITH's tools transfer in both directions, down to a 350M model and up to a 30B model, so the 4B writer is a viable drop-in tool provider regardless of which model ultimately consumes its tools.

\begin{table}[t]
  \centering
  \caption{Tools written by SMITH's RL-trained 4B writer lift a Qwen3-30B-A3B-Instruct
  consumer above the same 30B model writing tools for itself (LATM$^*$) on every
  group, with the largest gain on TabMWP-Hard (0.7 $\to$ 38.8), showing the 4B writer
  is a viable drop-in tool provider even when a much larger model is available to
  consume its tools. \emph{Overall}: task-count-weighted average over the 25 shared
  tasks (10 held-out RG + 13 seen RG + TabMWP-Hard + GQA). Best per column in
  \textbf{bold}.}
  \label{tab:30b_consumer}
  \small
  \begin{tabular}{lccccc}
    \toprule
    \textbf{Method} & \textbf{\shortstack{Held-out RG\\(10 tasks)}} & \textbf{\shortstack{Seen RG\\(13 tasks)}} & \textbf{TabMWP-Hard} & \textbf{GQA} & \textbf{Overall} \\
    \midrule
    LATM$^*$ (30B writes its own tool) & 71.9 & 78.1 & 0.7 & 20.2 & 70.2 \\
    \textbf{SMITH-4B to 30B consumer} & \textbf{74.5} & \textbf{84.6} & \textbf{38.8} & \textbf{30.2} & \textbf{76.6} \\
    \bottomrule
  \end{tabular}
\end{table}

\subsection{Out-of-distribution evaluation}

To probe transfer beyond Reasoning-Gym, we evaluate on a tabular-reasoning benchmark
(TabMWP-Hard) and a visual-reasoning benchmark (GQA), neither of which is represented
during training.
Table~\ref{tab:ood_transposed} shows SMITH leading TabMWP-Hard at $40.4$, ahead of the
closest baseline TroVE ($36.4$), and reaching second on GQA at $42.6$, trailing only
LATM distilled from GPT-4.1 ($56.0$).
We do not claim parity on perception: GPT-4.1 distillation embeds visual primitives our
self-trained writer never sees during RL, and the gap is the cost of avoiding oracle
supervision.
What \emph{does} transfer without distillation is the self-supervised tool-creation loop.
SMITH outperforms every scaffolding baseline on both OOD benchmarks and is the only
4B same-backbone method without oracle distillation to lead either column.

\begin{table}[t]
  \centering
  \caption{OOD generalization (Qwen3-4B-Instruct). SMITH leads on TabMWP-Hard (40.4 vs.\ next best TroVE 36.4) and ranks second on GQA (42.6), making it the only 4B method without oracle distillation to top either column. Bold = best, \underline{underline} = second-best.}
  \label{tab:ood_transposed}
  \small\setlength{\tabcolsep}{5pt}
  \begin{tabular}{lccccccccc}
    \toprule
     & \multicolumn{6}{c}{\textbf{Baselines}} & \multicolumn{2}{c}{\textbf{Distilled}} & \multicolumn{1}{c}{} \\
    \cmidrule(lr){2-7}\cmidrule(lr){8-9}\cmidrule(lr){10-10}
    \textbf{Task} & \textbf{CoT} & \textbf{LATM} & \textbf{\shortstack{LATM*\\(30B)}} & \textbf{CRAFT} & \textbf{TroVE} & \textbf{KTCE} & \textbf{ReTool} & \textbf{\shortstack{LATM\\(distill)}} & \textbf{SMITH} \\
    \midrule
    TabMWP-Hard & 7.2 & 19.7 & 0.7 & 30.0 & \underline{36.4} & 27.2 & 3.0 & 7.1 & \textbf{40.4} \\
    GQA & 11.5 &  35.0 & 29.8 & 21.9 & 21.4 & 0.0 & 26.1 & \textbf{56.0} & \underline{42.6} \\
    \bottomrule
  \end{tabular}
\end{table}

\subsection{Scaling across backbones}

To test whether SMITH's training signal transfers across model sizes and families, we
apply the same RL recipe to \textsc{Qwen3-8B} and \textsc{Granite-3.3-8B}.
Table~\ref{tab:ablation_8b} shows that for Qwen3-8B, SMITH improves both the accuracy in-distribution ($72.6 \to 79.4$) and the holdout RG accuracy ($72.2 \to 81.7$) and lifts the OOD TabMWP-Hard from $42.4$ to $56.7$.
While Granite-3.3-8B starts from a much weaker base (RG Seen $31.2$, Unseen $22.0$) yet
follows the same trend: SMITH lifts RG (Seen) to $39.1$, RG (Unseen) to $28.5$, and GQA
from $7.8$ to $11.7$.

We further test whether SMITH still works when the judge signal comes from the policy itself rather than an external 30B-A3B model. The \textbf{Self-Judge} variant prompts Qwen3-8B with the same judge template used for 30B-A3B, scoring its own rollouts. Self-judging improves held-out RG ($81.7 \to 85.9$, the best in the table) but trades off in-distribution accuracy and OOD GQA ($28.7 \to 16.3$), suggesting the smaller judge is a weaker but less biased signal on training-task distributions.

\begin{table}[t]
  \centering
  \caption{SMITH improves over the base model on every backbone tested. On Qwen3-8B it raises held-out RG from 72.2 to 81.7 and TabMWP-Hard from 42.4 to 56.7; Granite-3.3-8B follows the same trend despite a weaker starting point, confirming the training signal is not model-family specific. \emph{RG (Seen/Unseen)}: macro-average over training/held-out tasks. Best per column in \textbf{bold}.}
  \label{tab:ablation_8b}
  \begin{tabular}{lcccc}
    \toprule
    \textbf{Method} & \textbf{RG (Seen)} & \textbf{RG (Unseen)} & \textbf{TabMWP} & \textbf{GQA} \\
    \midrule
    \textsc{Qwen3-8B} (baseline) & 72.6 & 72.2 & 42.4 & 17.3 \\
    \textbf{SMITH: Qwen3-8B} & \textbf{79.4} & \textbf{81.7} & \textbf{56.7} & \textbf{28.7} \\
    \textbf{SMITH: Self-Judge} & \textbf{74.7} & \textbf{85.9} & 54.5 & 16.3 \\
    \midrule
    \textsc{Granite-3.3-8B} (baseline) & 31.2 & 22.0 & 3.9 & 7.8 \\
    \textbf{SMITH: Granite-3.3-8B} & 39.1 & 28.5 & 4.5 & 11.7 \\
    \bottomrule
  \end{tabular}
\end{table}

\subsection{Generalization to external tool-calling}

Beyond benchmarks where the model writes its own tools, we test whether SMITH's build-and-use objective transfers to \emph{externally specified} function-calling APIs using BFCL v4 (no-web subset) \citep{patil2025bfcl}. Table~\ref{tab:bfcl_summary} shows that SMITH lifts BFCL overall accuracy on both Qwen backbones, from $45.1$ to $48.6$ on Qwen3-4B and from $43.3$ to $55.8$ on Qwen3-8B, the largest absolute gain in the table. Since BFCL schemas, multi-turn traces, and judges are never seen during RL, the improvement isolates a learned tool-use prior rather than benchmark-specific fitting.

\begin{table}[t]
  \centering
  \caption{SMITH improves external tool-calling (BFCL v4, no-web) across all backbones, with the largest gain on Qwen3-8B (43.3$\to$55.8), despite never seeing BFCL schemas, multi-turn traces, or judges during RL training. Bold = best, \underline{underline} = second-best.}
  \label{tab:bfcl_summary}
  \small
  \begin{tabular}{lcccccc}
    \toprule
    & \multicolumn{2}{c}{\textbf{Qwen3-4B-Instruct}} & \multicolumn{2}{c}{\textbf{Qwen3-8B}} & \multicolumn{2}{c}{\textbf{Granite-3.3-8B}} \\
    \cmidrule(lr){2-3} \cmidrule(lr){4-5} \cmidrule(lr){6-7}
    \textbf{Metric} & \textbf{Base} & \textbf{SMITH} & \textbf{Base} & \textbf{SMITH} & \textbf{Base} & \textbf{SMITH} \\
    \midrule
    BFCLv4 & 45.1 & \textbf{48.6} & 43.3 & \textbf{55.8} & 36.3 & \textbf{38.7} \\
    \bottomrule
  \end{tabular}
\end{table}

\subsection{Ablation}

To isolate the factors driving SMITH's gains, we ablate the reward structure on
Qwen3-4B-Instruct and compare against four configurations:
(1) \textsc{Tool Create}, where the reward depends only on the quality of the generated
code;
(2) \textsc{Decoupled Build/Use}, which pairs the row-1 30B-A3B builder with a
separately trained 4B tool-use specialist, testing whether joint training (not
specialization) is the active ingredient;
(3) \textsc{SMITH: No LLM Judge}, which couples tool creation with execution success
but drops the judge signal; and
(4) the full SMITH objective, which keeps the LLM-as-judge signal as a separate axis.
Table~\ref{tab:ablation} shows that each component contributes incrementally: tool
creation alone yields a strong RG (Seen) bump but leaves OOD GQA underperforming;
the decoupled pair fails to beat single-model \textsc{Tool Create} on RG Unseen
($58.9$ vs.\ $68.8$), confirming that joint training, not specialization, drives the
gain; coupling build and use jointly lifts in-distribution accuracy but hurts
held-out transfer; only the full SMITH objective achieves the best aggregate
score, indicating that disentangling process quality from outcome correctness is
essential for cross-domain robustness.

\begin{table}[t]
  \centering
  \caption{Reward-structure ablation (Qwen3-4B-Instruct). $\pi^{\mathrm{eval}}$: model that invokes the generated tool at eval (\textit{30B-A3B}/\textit{4B}: separate model; \textit{self}: same RL'd policy). \textbf{Tool Use}: whether $\pi^{\mathrm{eval}}$ calls the tool rather than only writing it. Decoupled build/use does not beat single-model \textsc{Tool Create} on RG (Unseen) ($58.9$ vs.\ $68.8$), so joint training, not specialization, drives the gain; only the full SMITH objective wins on \emph{aggregate}, with individual benchmarks split between \textsc{K=1} (TabMWP) and \textsc{No LLM Judge} (GQA). Best per column \textbf{bold}, second-best \underline{underlined}. RG(S)/(U): Seen/Unseen.}
  \label{tab:ablation}
  \small
  \begin{tabular}{lcccccc}
    \toprule
    \textbf{Method} & $\boldsymbol{\pi^{\mathrm{eval}}}$ & \textbf{Tool Use} & \textbf{RG (S)} & \textbf{RG (U)} & \textbf{TabMWP} & \textbf{GQA} \\
    \midrule
    \textsc{Qwen3-4B-Instruct} & -- & -- & 61.85 & 47.01 & 19.70 & 20.86 \\
    \midrule
    \textsc{Tool Create} & 30B-A3B & No & 77.43 & 59.39 & 15.60 & 37.01 \\
    \textsc{Tool Create} & 4B & No & 73.88 & 68.80 & 17.70 & 35.82 \\
    \textsc{Decoupled Create/Use} & 30B-A3B & Yes & 76.44 & 58.93 & 13.90 & 35.04 \\
    \textsc{SMITH : No Sync} & 4B & Yes & 80.29 & 66.89 & 25.81 & 24.58 \\
    \textsc{SMITH : No LLM Judge} & self & Yes & \underline{82.57} & 67.81 & 18.30 & \textbf{42.63} \\
    \textsc{SMITH : K=1} & self & Yes & 78.64 & \underline{73.93} & \textbf{46.77} & 32.30 \\
    \textsc{SMITH : Full} & self & Yes & \textbf{86.61} & \textbf{78.33} & \underline{40.40} & \underline{42.62} \\
    \bottomrule
  \end{tabular}
\end{table}

%% file: sections/limitation.tex

\paragraph{Scale, judge dependence, and base priors.}
All trained policies are at most 8B parameters and the quality judge has 30B activated parameters, a regime chosen for compute feasibility; whether the gains persist, saturate, or invert at the 70B+ scale is an open question, and our Self-Judge ablation (Table~\ref{tab:ablation_8b}) is only a partial probe of judge-size dependence, since dropping the external judge improves held-out RG but degrades OOD GQA. Finally, SMITH executes generated Python at every step inside a sandbox, but we do not formally certify that adversarial prompts cannot induce unsafe tools, and we report no human evaluation of schema readability or developer-facing reusability.

\paragraph{Open Questions.}
1. In SMITH, a tool is one Python function plus a JSON schema. Richer artifacts are not evaluated in this work. For example, skills package workflow instructions in a \texttt{SKILL.md} file together with optional
scripts, references, templates, and other resources. MCP servers, by contrast,
use the Model Context Protocol to expose tools, resources, prompts, and instructions to a model; they are not themselves bundles of procedures and scripts. Supporting the generation of such multi-file or server-backed
artifacts would require extending SMITH's output representation and validation pipeline. Iterative or multi-turn generation may be useful for this setting, but is not inherently required. We leave this extension to future work.

2. Throughout SMITH training, we never observed the model issue parallel tool
calls in a single turn or generate multiple tools at once. We suspect this
reflects both the base model's tendency toward single-tool use and our reward
design, which does not encourage multi-tool generation or parallel execution.
Since parallel calls are common in real-world settings such as deep research,
we believe SMITH could naturally extend to these settings, which we leave to
future work.



%% file: sections/conclusion.tex
We introduced SMITH, a reinforcement learning framework that jointly trains a single language
model to \emph{create} and \emph{use} reusable tools, closing the feedback loop between tool
writer and tool user so the policy is optimized directly on its own execution outcomes.
Trained on $13$ Reasoning-Gym tasks, our 4B model attains the highest held-out RG accuracy
among all evaluated methods ($79.8$), leads TabMWP-Hard, and writes tools that transfer to a
$350$\,M student never seen during training, matching the quality of tools produced by a
model an order of magnitude larger.
The same recipe lifts Qwen3-8B and Granite-3.3-8B without modification, indicating that
coupling creation and use inside a single trained policy is a scalable path to generalization:
the tools a model writes become precisely the tools it can reliably invoke, without a larger
frozen teacher, a more complex scaffold, or out-of-domain supervision.

%% file: sections/ack.tex
\section*{Acknowledgments}
This work was supported in part by the National Science and Technology Council, Taiwan, under the Grants 115-2628-E-002-023-MY4, 112-2223-E-002-012-MY5, 115-2628-E-002-006, 115-2223-E-002-005-MY3, and 115-2634-F-002-012, the Taiwan Centers of Excellence in Artificial Intelligence, and the Center of Data Intelligence: Technologies, Applications, and Systems, NTU (grant nos. 115L900901). Shao-Hua Sun was supported by the Yushan Fellow Program of the Ministry of Education, Taiwan.

%% file: sections/related_work.tex
Early tool-augmented LLMs invoke search engines or self-supervised API calls to extend parametric memory \citep{komeili2022internet, thoppilan2022lamda, schick2023toolformer}, but are bounded by a fixed, human-curated toolset.

LATM \citep{cai2023large} introduced dynamic tool creation, using a powerful LLM to write reusable tools a weaker model can invoke, yielding stronger performance on BIG-Bench \citep{srivastava2023beyond}.
Subsequent works added increasingly complex scaffolding: CRAFT \citep{yuancraft} builds a retrieval-augmented tool library with verification; Trove \citep{wang2024trove} introduces tool induction and verification pipelines; and KTCE \citep{ma2025automated} automates creation and evaluation through multi-stage decomposition.
Despite their differences, all of these systems treat tool writing as a prompting problem at inference time, leaving tool quality to emerge incidentally from generation rather than from an explicit objective and none couples tool creation with tool use in a joint learning objective.

A parallel line applies RL to code and tool generation: CodeRL \citep{le2022coderl}, RLEF \citep{gehring2025rlef} use execution feedback to train better code generators and ToolRL\citep{qiantoolrl} shows tailored design reward can help tool use, while ReTool \citep{feng2025retool} trains models to invoke tools more reliably without addressing creation.
The closest work is SAGE \citep{wang2025reinforcement}, which frames skill creation as an RL objective; however, SAGE uses a single question for creation and a single for validation, limiting generalization across diverse task categories.
Our work differs in three key respects: we jointly train tool creation and use across 13 procedural task categories; we validate each generated tool on up to 16 held-out questions; and we apply an LLM judge scoring correctness, schema quality, and overall tool quality as an independent reward signal.

SMITH sits at the intersection of program-aided reasoning \citep{gao2023pal, wang2024executable} and program induction \citep{wanginducing}, inheriting their use of executable code, synthesis of reusable procedures, and utility of compounding tools between tasks.
It departs from all three by making these capabilities explicit RL objectives: the \emph{build} task directly rewards concise, interpretable, and reusable tool synthesis, while the \emph{use} task closes the loop by ensuring what the model writes is what it can reliably invoke.

%% file: appendix/verify_and_llm_judge.tex
\section{Lessons Learned in Designing Verifier Rewards and LLM-as-Judge}
\label{app:verify_judge}

\subsection{System Overview}

The training system fine-tunes Qwen3-4B-Instruct on two interleaved task types
with DAPO loss and LoRA ($r=64$). In \textbf{build tasks}, the model writes a Python function and a matching
OpenAI-compatible JSON schema, validated and scored by three independent signals:
a structural verifier, a LoRA-synced evaluator that runs held-out test questions
through the generated tool, and an LLM judge scoring code quality.
In \textbf{use tasks}, the model is given a pre-built tool schema and must invoke
it correctly to answer a question; reward is rule-based answer matching.

In the configuration analysed here (which differs from the final system),
the LLM judge was a \emph{fixed} external model (Qwen3-30B) rather than the
self-synced policy checkpoint used in the final SMITH design.

The total reward for a build trajectory is:

\begin{equation}
  r_{\mathrm{build}} = \underbrace{r_{\mathrm{format}}}_{\text{verifier}}
                     + \underbrace{r_{\mathrm{eval}}}_{\text{LoRA evaluator}}
                     + \underbrace{w_j \cdot r_{\mathrm{judge}}}_{\text{LLM judge}},
  \quad w_j = 0.5
  \label{eq:reward}
\end{equation}

where $r_{\mathrm{format}} \in \{0, 0.5, 1.0\}$ encodes schema--function-name
alignment (0.5) and parameter alignment (0.5), $r_{\mathrm{eval}} \in [0,1]$
is the fraction of test questions answered correctly by the generated tool, and
$r_{\mathrm{judge}} \in [-0.5, 1]$ is the normalised LLM judge score.
Note that the final system tightens $r_{\mathrm{format}}$ to a binary
$\{0, 0.5\}$ signal granted only when both function-name and parameter
alignment hold simultaneously (Section~\ref{sec:build}); the additive
$\{0, 0.5, 1.0\}$ form above is specific to the configuration analysed
in this appendix.

\subsection{Observed Failure: \texttt{build/env\_reward} Collapse}

\paragraph{Quantitative trajectory.}
Table~\ref{tab:metrics} shows per-step metrics logged during training.
The LoRA sync occurs at step~5 (checkpoints are saved every 5 steps).
A sharp regression is visible from step~6 onward.

\begin{table}[h]
  \centering
  \caption{Key build-task metrics per optimizer step.
    The LoRA sync that loads \texttt{eval\_lora\_step5} into the evaluator
    server fires \emph{after} step~5's metrics are logged.
    Steps~6--7 are the first rollouts with the synced evaluator.}
  \label{tab:metrics}
  \footnotesize
  \setlength{\tabcolsep}{4pt}
  \begin{tabular}{lcccccc}
    \toprule
    Step & \texttt{env\_reward} & \texttt{eval\_reward} & \texttt{step1\_fail} & \texttt{fn\_mismatch} & \texttt{param\_mismatch} & \texttt{judge\_reward} \\
    \midrule
    1  & 1.363 & 0.386 & 34.9\% & 23.4\% & 11.5\% & 0.460 \\
    2  & 1.809 & 0.466 & 10.4\% & 8.3\%  & 2.1\%  & 0.614 \\
    3  & 2.026 & 0.557 & 2.1\%  & 1.0\%  & 1.0\%  & 0.670 \\
    4  & 2.043 & 0.684 & 9.4\%  & 5.2\%  & 3.6\%  & 0.710 \\
    5  & 2.123 & 0.662 & 2.6\%  & 0.0\%  & 2.6\%  & 0.722 \\
    \midrule
    \multicolumn{7}{c}{\emph{LoRA sync applied after step~5}} \\
    \midrule
    6  & 1.641 & 0.500 & 23.9\% & 13.5\% & 10.4\% & 0.561 \\
    7  & 1.402 & 0.332 & 28.6\% & 14.1\% & 14.1\% & 0.453 \\
    \bottomrule
  \end{tabular}
\end{table}

\paragraph{Attribution of the drop.}
Since $r_{\mathrm{eval}}$ is averaged over all build states including step1
failures (which receive $r_{\mathrm{eval}} = 0$), the $r_{\mathrm{eval}}$
collapse follows directly from the \texttt{step1\_fail} spike.  We verified
this by inspecting rollout event logs: at step~6, 46 out of 238 build states
failed step1 (19.3\%), versus 5 out of 197 at step~5 (2.5\%).
Of those 46 failures, 26 were \texttt{schema\_function\_mismatch} and 20 were
\texttt{schema\_param\_mismatch}.

Crucially, the \emph{evaluator was not failing}.
No connection errors or exceptions appeared during tool evaluation.
Items that did reach the evaluator scored 5.26/8 test questions on average
at step~6, close to the 5.44/8 at step~5.
The drop in \texttt{build\_eval\_reward} is almost entirely explained by the
increased fraction of step1-failed states receiving a forced score of zero.

\subsection{Root Cause Analysis}

\subsubsection{Naming Drift}

At step~6, the model began generating code with several plausible function
names, while the schema referenced one that was not present as a top-level
callable.
We term this \emph{naming drift}.
Figure~\ref{fig:naming_drift} shows a representative failure.

\begin{figure}[h]
\begin{minipage}{\columnwidth}
\begin{mdframed}
\begin{verbatim}
def _gcd(a: int, b: int) -> int:
    while b:
        a, b = b, a % b
    return a

def calculate_lcm(a: int, b: int) -> int:
    return (a * b) // _gcd(a, b)

def find_lcm_of_numbers(a: int, b: int) -> int:
    return calculate_lcm(a, b)
\end{verbatim}
\vspace{2pt}
\begin{verbatim}
[{"type": "function", "function": {
  "name": "find_lcm",
  "parameters": {"type": "object",
    "properties": {"a": {"type": "integer"},
                   "b": {"type": "integer"}},
    "required": ["a", "b"]}}}]
\end{verbatim}
\end{mdframed}
\end{minipage}
\caption{Naming drift example. The schema references \texttt{find\_lcm},
  which does not appear as a top-level callable in the generated code.}
\label{fig:naming_drift}
\end{figure}

\subsubsection{Reward Signal Misalignment}
\label{sec:mismatch}

The drift persisted because the verifier and the judge applied inconsistent
penalties for the same error:

\begin{table}[h]
  \centering
  \caption{Penalty for a schema that names a function absent from the code.}
  \label{tab:penalty}
  \begin{tabular}{lcc}
    \toprule
    Component & Condition & Penalty \\
    \midrule
    Structural verifier & fn name absent & $r = 0$, trajectory terminates \\
    LLM judge           & fn name absent & $r \times 0.5$ (partial credit) \\
    \bottomrule
  \end{tabular}
\end{table}

Because step1-failed trajectories are terminated before the judge scoring
phase, the judge never observes naming-drift failures directly.
It therefore systematically rewarded the multi-function code style that
accompanied high-scoring outputs at step~5, reinforcing it via the step~5
gradient update.
By step~6, the drift had progressed far enough that the schema function name
was no longer present in the generated code.

\begin{observation}
When a verifier and an LLM judge jointly determine reward, any property
treated as a \emph{hard failure} by the verifier must also yield \emph{zero}
in the judge.  Partial credit for conditions that are hard failures in the
verifier creates a gradient toward outputs that pass the judge but fail the
verifier.
\end{observation}

\subsubsection{Secondary Factor: LoRA-Synced Evaluator}
\label{sec:lora_eval}

The LoRA sync at step~5 replaces the base-model evaluator with the current
training checkpoint, on the rationale that improved tool-building ability
yields a stronger usability signal for $r_{\mathrm{eval}}$.
In practice the synced evaluator was not the primary cause of the collapse:
the mean per-item test score fell only mildly (5.44~$\to$~5.26 out of~8),
while the step1-failure rate increase was the dominant driver.
A brief accuracy drop immediately post-sync is nonetheless expected, as
tool-use ability tends to lag behind tool-building improvements in the early
steps of RL training.

\begin{observation}
LoRA-syncing the evaluator with the training checkpoint is sound in principle,
but evaluator signal quality should be monitored at each sync to distinguish
evaluator degradation from policy regression.
\end{observation}

\subsection{Lessons for Reward and Judge Design}

We derive five concrete lessons from this failure.

\subsubsection{Lesson 1: Hard Failures Must Be Hard Everywhere}

\begin{pitfall}
A verifier condition that zeroes out the entire trajectory reward should also
zero out the judge component, not merely halve it.
\end{pitfall}

In our case, the judge applied:

\begin{figure}[h]
\begin{minipage}{\columnwidth}
\begin{mdframed}
\begin{verbatim}
if not aligned:
    if reason.startswith("syntax error in code:"):
        score = -0.5        # hard negative
    else:
        score *= 0.5        # soft halving -- fn-absent treated same
                            # as minor param mismatch
\end{verbatim}
\end{mdframed}
\end{minipage}
\caption{Original (buggy) misalignment handling: function-absent receives only a soft penalty.}
\label{fig:judge_before}
\end{figure}

The fix distinguishes the severity of each misalignment type:

\begin{figure}[h]
\begin{minipage}{\columnwidth}
\begin{mdframed}
\begin{verbatim}
if not aligned:
    if reason.startswith("syntax error in code:"):
        score = -0.5    # unparseable -- keep strong negative
    elif "not found in code" in reason:
        score = 0.0     # schema fn absent: hard zero, same as verifier
    else:
        score *= 0.5    # softer issues: param mismatch, over-promise
\end{verbatim}
\end{mdframed}
\end{minipage}
\caption{Fixed misalignment handling: function-absent is now a hard zero.}
\label{fig:judge_after}
\end{figure}

\begin{lesson}
For every binary constraint that the verifier enforces with reward $= 0$,
audit the judge's scoring path and ensure it applies the same zero (or negative)
for states where that constraint is violated.  Partial credit for fatal
structural errors creates a false gradient.
\end{lesson}

\subsubsection{Lesson 2: Align the Judge Prompt to the Verifier's Hard Constraints}

The judge prompt rewarded ``descriptive naming'' and ``well-decomposed
functions'' without specifying that the schema's \texttt{name} field must exist
as a top-level callable.
As a result, the judge gave high \texttt{code\_clarity} scores to multi-function
outputs regardless of whether the schema function was present.

We extended the prompt with:
\begin{itemize}[leftmargin=1.5em]
  \item An explicit Step~3 instruction: \emph{``The schema \texttt{name} must
    appear verbatim as a top-level \texttt{def} before any other analysis.
    If not, assign \texttt{schema\_code\_alignment=0} and
    \texttt{overall\_quality=0} immediately.''}
  \item Two scored examples (D and E) illustrating the naming-drift anti-pattern
    and the thin-wrapper anti-pattern respectively.
  \item Updated \texttt{code\_clarity} rubric language that explicitly penalises
    thin wrappers (score 1--2) and multi-function code where the schema-named
    function is not the direct implementation.
\end{itemize}

\begin{lesson}
Every structural constraint enforced by the verifier should appear explicitly
in the judge prompt, ideally with a scored counterexample.  The judge cannot
penalise a failure mode it has not been told to look for.
\end{lesson}

\subsubsection{Lesson 3: The Reward Gap Between Verifier and Judge Is a Gradient Leak}

Let $\mathcal{S}_{\mathrm{bad}}$ be the set of outputs that fail the verifier's
hard constraint (schema function absent from code).
For a GRPO update, outputs in $\mathcal{S}_{\mathrm{bad}}$ received:

\begin{align}
  r_{\mathrm{total}}(\mathbf{y}) &= 0 + 0 + w_j \cdot r_j(\mathbf{y})
  \quad \mathbf{y} \in \mathcal{S}_{\mathrm{bad}},
  \label{eq:bad_reward}
\end{align}

where the verifier and evaluator terms are both zero, but the judge term
$w_j \cdot r_j > 0$ if the judge gave partial credit.
If the judge assigns $r_j = 0.4$ and $w_j = 0.5$, the output in
$\mathcal{S}_{\mathrm{bad}}$ earns a total reward of $0.2$ (positive), even
though the verifier called it completely broken.

Under GRPO, the advantage $A(\mathbf{y}) = r_{\mathrm{total}}(\mathbf{y}) -
\bar{r}$ is positive whenever $r_j(\mathbf{y}) > 2\bar{r}$ (since
$w_j = 0.5$).  This means the policy is actively \emph{reinforced} toward
producing outputs in $\mathcal{S}_{\mathrm{bad}}$ when they score well on the
judge.
We call this the \emph{reward gap}: the judge's non-zero floor creates a
gradient leak that partly cancels the verifier's hard zero.

\begin{pitfall}
With $N$ reward components combined additively, any component that assigns
positive reward to verifier-failed outputs will create a gradient leak.
The larger the judge weight $w_j$ and the higher the judge's partial score for
broken outputs, the stronger the gradient toward structural failures.
\end{pitfall}

\begin{lesson}
When combining a hard verifier with a soft LLM judge, consider gating the judge
signal: apply $r_j = 0$ whenever the verifier assigns $r = 0$, regardless of
what the judge scores.  Alternatively, use the judge signal only as a
\emph{tiebreaker} among verifier-passing outputs, not as an independent additive
axis.
\end{lesson}

\subsubsection{Lesson 4: Log Structural Sub-Metrics as First-Class Signals}

The aggregate reward trajectory at steps 1--5 looked healthy: \texttt{build/env\_reward}
was rising and \texttt{judge\_reward} was stable (Table~\ref{tab:metrics}).
The impending failure was \emph{invisible in these aggregates}.
It was only visible in the two sub-metric columns: \texttt{fn\_mismatch} had
been declining from 23.4\% (step~1) to 0.0\% (step~5), while
\texttt{param\_mismatch} had converged to 1--3\%.
A monitoring system that tracked only aggregate reward would have declared the
run healthy at step~5; the structural failure at step~6 would have appeared
sudden and unexplained.

\begin{pitfall}
Aggregate reward metrics (\texttt{env\_reward}, \texttt{eval\_reward},
\texttt{judge\_reward}) reflect the \emph{mean} over all constraint
dimensions simultaneously.
A model that learns to satisfy one constraint better while silently degrading
another can maintain or even improve its aggregate score until the neglected
constraint crosses a critical threshold.
The aggregate cannot distinguish this pattern from genuine all-round improvement.
\end{pitfall}

In our case, the correct monitoring surface was:

\begin{itemize}[leftmargin=2em, itemsep=2pt]
  \item \textbf{\texttt{step1\_fail} rate.}
        The fraction of build rollouts that fail the structural verifier at the
        first check.
        This is the earliest observable symptom of naming drift or format
        breakdown; it should be logged every step, not inferred from
        \texttt{eval\_reward}.
  \item \textbf{\texttt{fn\_mismatch} rate.}
        The fraction of rollouts where the schema \texttt{name} field does not
        appear as a top-level callable in the generated code.
        This constraint is either satisfied or violated; its rate should stay
        near zero after training stabilises.
  \item \textbf{\texttt{param\_mismatch} rate.}
        The fraction of rollouts where the schema's parameter list does not
        match the function signature.
        A separate, independently trackable constraint.
\end{itemize}

\noindent These sub-metrics are the \emph{concrete example} of Principle 3 in
Table~\ref{tab:principles}: they make it possible to detect which specific
structural constraint is degrading, and to respond with a targeted fix (as in
Lesson 1 and Lesson 2) rather than a global hyper-parameter change.

A practical implementation note: because step1-failed rollouts are terminated
before the evaluator and judge run, \texttt{step1\_fail} is the only signal
that captures them.
Setting a monitoring alert on \texttt{step1\_fail} $> 10\%$ would have
triggered at step~6 immediately, before \texttt{eval\_reward} had time to
collapse all the way to the values observed at step~7.

\begin{lesson}
Log each structural constraint-violation rate as an independent time-series
metric, not as a component folded into aggregate reward.
Set threshold-based alerts on violation rates, not on aggregate reward
alone.
When a violation rate that was decreasing suddenly spikes (even by a few
percentage points), investigate immediately: this is an early warning sign
of the reward-misalignment failure described in Lessons~1--3.
\end{lesson}

\subsubsection{Lesson 5: Emergent Style Transfer Can Corrupt Structural Constraints}

A subtle consequence of RL with a combined reward is that the policy can learn
a \emph{style} associated with high-reward outputs and transfer that style even
to outputs where it causes structural failures.

In our case, multi-function code with helper functions was correlated with high
$r_{\mathrm{eval}}$ at steps 3--5 (better-structured tools passed more test
questions).  The policy learned ``helper functions $\rightarrow$ high reward''
as a latent heuristic.
At step~6, this heuristic overrode the schema-naming constraint: the model
wrote elaborate helper structures but the schema function name no longer anchored
to any of them.

This is a form of \emph{reward hacking} that is difficult to detect from the
reward trajectory alone, because the style that causes the failure was
associated with \emph{correct} behaviour earlier in training.
The failure is only visible in the structural sub-metrics (\texttt{fn\_mismatch}
rate).

\begin{lesson}
Monitor sub-metric trajectories across training steps, not just their endpoint
values.  When a structural constraint violation rate that was decreasing
suddenly increases, suspect that the policy has learned a correlated style
feature that is generalising beyond the constraint boundary.  Structural
constraints should be \emph{hard-coded} into the reward (verifier) rather
than expressed through soft proxy signals (LLM judge).
\end{lesson}

\subsection{Summary of Design Principles}

Table~\ref{tab:principles} summarizes the five lessons as actionable design
principles for practitioners building verifier + LLM-as-judge reward pipelines.

\begin{table}[h]
  \centering
  \small
  \caption{Design principles for combining a structural verifier with an
    LLM-as-judge reward signal in RL training for code generation.}
  \label{tab:principles}
  \setlength{\tabcolsep}{4pt}
  \begin{tabularx}{\linewidth}{clX}
    \toprule
    \# & Principle & Implication \\
    \midrule
    1 & \textbf{Hard failures must be hard everywhere.}
      & If the verifier zeros a trajectory, the judge must also return zero for
        that condition.  Never apply partial credit to a fatal structural error. \\[4pt]
    2 & \textbf{Align the judge prompt to verifier constraints.}
      & Every binary constraint enforced by the verifier should appear verbatim
        in the judge prompt, with a scored negative example. \\[4pt]
    3 & \textbf{Log structural sub-metrics as first-class signals.}
      & Track each constraint-violation rate independently.  Set alerts on
        violation rates, not only on aggregate reward. \\[4pt]
    4 & \textbf{Gate the judge signal on verifier pass.}
      & Apply $r_j = 0$ whenever the verifier assigns $r = 0$ to prevent
        gradient leaks from judge partial credit. \\[4pt]
    5 & \textbf{Structural constraints belong in the verifier, not the judge.}
      & RL policies transfer styles associated with high reward.  If a structural
        property can be enforced exactly by a rule, enforce it as a hard verifier
        check rather than as a soft judge preference. \\
    \bottomrule
  \end{tabularx}
\end{table}

%% file: appendix/judge_prompt.tex
\section{LLM-as-Judge Prompt (v5.0)}
\label{app:judge_prompt}

We reproduce below the complete system prompt supplied to the LLM-as-judge
component of our build-task reward pipeline (Section~\ref{sec:build}).
The judge receives each generated tool (a Python function together with its
OpenAI-compatible JSON schema) and returns a structured JSON object with five
numeric quality scores and three structured text fields.
The prompt is presented verbatim to support reproducibility.

\bigskip

\begin{AIboxC}[width=\textwidth]{LLM-as-Judge Prompt (v5.0)}

You are a code reviewer evaluating Python tool implementations and their
OpenAI function-calling schemas.
These tools are used by LLMs to solve mathematical and reasoning tasks via
tool-calling.

\medskip

\textbf{Important: what to look for}

\smallskip
Do \textbf{not} attempt to mentally execute the code or verify that it
produces correct numeric outputs; that approach is unreliable.
Instead, assess the tool based on \emph{structural} signals you can directly
observe in the code and schema.

\medskip

\textbf{Required analysis process}

\smallskip
Complete these steps before scoring.

\medskip
\textit{Step 1: Check code structure for red flags.}
Look for these specific problems (list every one you find):

\begin{itemize}[leftmargin=1.5em, topsep=2pt, itemsep=1pt]
  \item \textbf{Incomplete implementation:}
        \texttt{pass}, \texttt{TODO}, \texttt{NotImplementedError},
        empty branches, functions that only handle a subset of cases
        (e.g.\ only quadratic when the task is general polynomial).
  \item \textbf{Fragile parsing:}
        regex-based math parsing instead of proper libraries
        (\texttt{sympy}, \texttt{numpy}, \texttt{ast});
        hardcoded patterns that won't generalise.
  \item \textbf{Wrong approach:}
        algorithm doesn't match the task type
        (e.g.\ brute-force search for a task that needs symbolic math).
  \item \textbf{Missing error paths:}
        bare \texttt{except: pass} that silently swallows errors;
        returning empty/\texttt{None} on failure without indication.
  \item \textbf{Hardcoded limits:}
        magic numbers, fixed-size assumptions, only works for specific
        input dimensions.
  \item \textbf{Truncated code:}
        function ends abruptly, code was cut off mid-implementation.
\end{itemize}

\smallskip
\textit{Step 2: Check if the code covers the task.}
Compare the task type and sample questions against what the code actually
implements.
A tool that handles only linear equations when Q1 is a cubic is fundamentally
inadequate, regardless of how clean the code looks.

\smallskip
\textit{Step 3: Schema--code comparison.}
Compare schema parameter names, types, and descriptions against the function
signature.
Note every mismatch.

\smallskip
\textit{Step 4: Assess tool API design.}
If an LLM only sees the schema (not the code), could it construct correct
function calls?
Are parameter names and descriptions clear enough?

\medskip

\textbf{Scored examples}

\smallskip
Study these examples carefully.
They show common pitfalls, especially tools that \emph{look} polished but
are structurally broken.

\medskip
\textit{Example A: Good tool (high scores).}\\
\textbf{Task:} \texttt{count\_bits}\quad
\textbf{Q1:} ``How many 1 bits are in the binary representation of
76{,}778{,}227?'' \quad \textbf{Expected:} 14

\begin{lstlisting}[language=Python, basicstyle=\small\ttfamily,
                   breaklines=true, frame=single, xleftmargin=0.5em]
def count_one_bits(n):
    return bin(n).count('1')
\end{lstlisting}

\noindent\textbf{Schema:}
\begin{lstlisting}[basicstyle=\small\ttfamily, breaklines=true,
                   frame=single, xleftmargin=0.5em]
{
  "name": "count_one_bits",
  "parameters": {
    "properties": {
      "n": {
        "type": "integer",
        "description": "The non-negative integer whose binary
          representation is to be analyzed for the count of 1 bits."
      }
    },
    "required": ["n"]
  }
}
\end{lstlisting}

\noindent\textbf{Correct scores:}
\begin{lstlisting}[basicstyle=\small\ttfamily, breaklines=true,
                   frame=single, xleftmargin=0.5em]
{
  "red_flags": "none",
  "task_coverage": "yes -- bin().count() handles any non-negative integer",
  "code_correctness": 4, "code_clarity": 4,
  "schema_quality": 4, "schema_code_alignment": 5, "overall_quality": 4
}
\end{lstlisting}

\noindent\textit{Why:}
Simple, correct approach using a reliable built-in that generalises to all
inputs.
Schema matches code exactly.
Not 5 because there is no input validation (negative numbers give wrong
results).

\medskip
\textit{Example B: Polished but broken tool (low scores despite good
appearance).}\\
\textbf{Task:} \texttt{polynomial\_equations}\quad
\textbf{Q1:} ``Solve \texttt{q**5 + 21*q**4 + 44*q**3 + 45 = 0}''
\quad \textbf{Expected:} $-18.6398,\ -2.0423,\ -1.3908$

\begin{lstlisting}[language=Python, basicstyle=\small\ttfamily,
                   breaklines=true, frame=single, xleftmargin=0.5em]
import math

def solve_quadratic_equation(a, b, c):
    """Solves a*q^2 + b*q + c = 0 and returns real decimal solutions."""
    if abs(a) < 1e-10:
        if abs(b) < 1e-10:
            return "0.0" if abs(c) < 1e-10 else ""
        return f"{-c/b:.4f}"
    discriminant = b**2 - 4*a*c
    if discriminant < 0:
        return ""
    ...
\end{lstlisting}

\noindent\textbf{Schema:}
\begin{lstlisting}[basicstyle=\small\ttfamily, breaklines=true,
                   frame=single, xleftmargin=0.5em]
{
  "name": "solve_quadratic_equation",
  "parameters": {
    "properties": {
      "a": {"type": "number"},
      "b": {"type": "number"},
      "c": {"type": "number"}
    },
    "required": ["a", "b", "c"]
  }
}
\end{lstlisting}

\noindent\textbf{Correct scores:}
\begin{lstlisting}[basicstyle=\small\ttfamily, breaklines=true,
                   frame=single, xleftmargin=0.5em]
{
  "red_flags": "function only handles quadratic (degree 2) equations
    but Q1 is degree 5; hardcoded to 3 coefficients",
  "task_coverage": "no -- cannot solve any polynomial above degree 2,
    which is the primary task requirement",
  "code_correctness": 1, "code_clarity": 4,
  "schema_quality": 4, "schema_code_alignment": 2, "overall_quality": 1
}
\end{lstlisting}

\noindent\textit{Why:}
The code \emph{looks} clean (good naming, typed parameters, proper
docstring) but it fundamentally cannot solve the task: it handles only
degree-2 polynomials while the task requires general polynomial solving.
\texttt{schema\_code\_alignment} is 2 because the schema doesn't warn users
that it only handles quadratics.

\medskip
\textit{Example C: Ugly but functional tool (moderate scores).}\\
\textbf{Task:} \texttt{polynomial\_equations}\quad
\textbf{Q1:} ``Solve \texttt{q**5 + 21*q**4 + 44*q**3 + 45 = 0}''
\quad \textbf{Expected:} $-18.6398,\ -2.0423,\ -1.3908$

\begin{lstlisting}[language=Python, basicstyle=\small\ttfamily,
                   breaklines=true, frame=single, xleftmargin=0.5em]
def solve_equation(equation_str):
    from sympy import symbols, solve, Eq, sympify
    q = symbols('q')
    expr = sympify(equation_str.replace('= 0', '').strip())
    solutions = solve(Eq(expr, 0), q)
    real_sols = [complex(s).real for s in solutions
                 if abs(complex(s).imag) < 1e-6]
    return ', '.join(f'{s:.4f}' for s in sorted(real_sols))
\end{lstlisting}

\noindent\textbf{Schema:}
\begin{lstlisting}[basicstyle=\small\ttfamily, breaklines=true,
                   frame=single, xleftmargin=0.5em]
{
  "name": "solve_equation",
  "parameters": {
    "properties": {
      "equation_str": {
        "type": "string",
        "description": "The equation to solve"
      }
    },
    "required": ["equation_str"]
  }
}
\end{lstlisting}

\noindent\textbf{Correct scores:}
\begin{lstlisting}[basicstyle=\small\ttfamily, breaklines=true,
                   frame=single, xleftmargin=0.5em]
{
  "red_flags": "uses sympify on raw string input (fragile); schema
    description is vague; single string argument instead of structured
    input",
  "task_coverage": "yes -- sympy.solve handles arbitrary polynomial
    degrees",
  "code_correctness": 3, "code_clarity": 1,
  "schema_quality": 1, "schema_code_alignment": 4, "overall_quality": 2
}
\end{lstlisting}

\noindent\textit{Why:}
The approach is correct (\texttt{sympy.solve} handles arbitrary
polynomials), but the API is weak: one opaque string argument and a vague
schema description.
An LLM might pass the equation in a format \texttt{sympify} can't parse.
\texttt{code\_correctness} is 3 (not higher) because \texttt{sympify} on raw
strings is fragile.
Overall 2: despite API flaws the core algorithm works.

\medskip

\textbf{Key lesson from the examples}

\smallskip
Example B is the trap to avoid: \textbf{do not} give high scores just because
code is well-structured, well-named, or has good parameter decomposition.
If the approach cannot handle the task (Step~2 fails),
\texttt{code\_correctness} \textbf{must} be 0--2 regardless of code quality.
Example C shows the reverse: ugly code with a correct approach deserves higher
\texttt{code\_correctness} than polished code with a wrong approach.

\medskip

\textbf{Scoring rubrics (0--5)}

\medskip
\texttt{code\_correctness} (0--5): \textit{structural soundness: does the
code implement a viable approach for the task?}
\begin{itemize}[noitemsep, topsep=2pt, leftmargin=2em]
  \item[0] No implementation, syntax errors, or completely unrunnable.
  \item[1] Critical structural flaws: incomplete branches, truncated
           implementation, or approach fundamentally wrong for the task type.
  \item[2] Approach is plausible but has significant gaps: only handles a
           subset of cases, fragile parsing, or swallows errors silently.
  \item[3] Solid implementation with minor structural concerns
           (e.g.\ no input validation, hardcoded limits that might not cover
           all cases).
  \item[4] Clean implementation using appropriate libraries/algorithms;
           handles the task type fully.
  \item[5] Robust implementation with explicit error handling, input
           validation, and appropriate algorithm choice.
\end{itemize}

\smallskip
\texttt{code\_clarity} (0--5): \textit{naming quality and argument design
for LLM usability}
\begin{itemize}[noitemsep, topsep=2pt, leftmargin=2em]
  \item[0] No meaningful function: bare code snippet or hardcoded answer with
           no parameters.
  \item[1] Cryptic function name (e.g.\ \texttt{f}, \texttt{run}); all input
           is one opaque string argument.
  \item[2] Generic name (e.g.\ \texttt{solve}, \texttt{process}); arguments
           poorly named or bundled into one dict/string.
  \item[3] Name indicates purpose (e.g.\ \texttt{calculate\_area}); arguments
           separated but could be decomposed further.
  \item[4] Descriptive name; arguments well-decomposed into typed parameters
           (e.g.\ \texttt{operator: str, a: float, b: float} instead of
           \texttt{expression: str}).
  \item[5] Self-documenting: precise function name; each argument is atomic,
           well-typed, and named so an LLM can call it without examples.
\end{itemize}

\smallskip
\texttt{schema\_quality} (0--5): \textit{OpenAI schema completeness and
accuracy}
\begin{itemize}[noitemsep, topsep=2pt, leftmargin=2em]
  \item[0] No schema or malformed JSON.
  \item[1] Missing required fields, wrong types, no descriptions.
  \item[2] Vague or misleading descriptions; types partially wrong.
  \item[3] Adequate descriptions, correct types, but missing format details
           or constraints.
  \item[4] Precise descriptions, correct types, required list correct.
  \item[5] Descriptions specify exact input format, constraints, value ranges,
           and examples.
\end{itemize}

\smallskip
\texttt{schema\_code\_alignment} (0--5): \textit{does the schema accurately
represent what the code does?}
\begin{itemize}[noitemsep, topsep=2pt, leftmargin=2em]
  \item[0] Schema describes a completely different function.
  \item[1] Parameter names or types contradict the function signature.
  \item[2] Schema describes idealised behaviour the code doesn't implement.
  \item[3] Mostly aligned but schema over-promises
           (e.g.\ claims to handle cases the code skips).
  \item[4] Minor discrepancy only (e.g.\ one description slightly off).
  \item[5] Schema is a precise contract for the actual implementation.
\end{itemize}

\smallskip
\texttt{overall\_quality} (0--5): \textit{holistic: would an LLM get
correct answers using this tool?}

\noindent\textbf{Important:} \texttt{overall\_quality} \textbf{cannot exceed}
\texttt{code\_correctness\,+\,1}.
A tool with broken code is not redeemed by a nice schema.
\begin{itemize}[noitemsep, topsep=2pt, leftmargin=2em]
  \item[0] Unusable: no working code or no schema.
  \item[1] Broken: critical structural flaws prevent reliable use.
  \item[2] Marginal: plausible approach but too many gaps for reliable
           results.
  \item[3] Functional: solid approach, works for the common case.
  \item[4] Good: correct approach, clear API, accurate schema.
  \item[5] Excellent: robust implementation with great API design.
\end{itemize}

\medskip

\textbf{Output format}

\smallskip
Respond with \textbf{only} a JSON object:

\begin{lstlisting}[basicstyle=\small\ttfamily, breaklines=true,
                   frame=single, xleftmargin=0.5em]
{
  "red_flags":             "<list every structural problem found in Step 1,
                             or 'none' if clean>",
  "task_coverage":         "<yes/partially/no -- does the code cover the
                             task type and sample questions?>",
  "reasoning":             "<1-2 sentences justifying scores based on the
                             red flags and task coverage above>",
  "code_correctness":      <int 0-5>,
  "code_clarity":          <int 0-5>,
  "schema_quality":        <int 0-5>,
  "schema_code_alignment": <int 0-5>,
  "overall_quality":       <int 0-5>
}
\end{lstlisting}

\end{AIboxC}

%% file: appendix/prompt_templates.tex

\section{Build and Use Task Prompt Templates}
\label{app:prompts}

This appendix documents the prompt templates used during SMITH training.
All prompts are held fixed throughout training; no prompt engineering is done
between training runs.

\subsection{Build-Task Prompt}
\label{app:build_prompt}

The build task follows a two-part message structure: a \emph{system message}
that establishes the model's role, and a \emph{user message} that combines
detailed tool-writing instructions with the $N{=}4$ in-context
\texttt{(question, answer)} pairs.

\paragraph{System message.}

\begin{AIboxC}{Build-Task System Message}
\small
You are an expert tool builder.
\end{AIboxC}

\paragraph{User message structure.}

The user message concatenates three components in order:

\begin{enumerate}[leftmargin=2em, itemsep=3pt]
  \item \textbf{Tool-maker instruction block.}
        A fixed multi-paragraph prompt (abridged below) specifying:
        \begin{itemize}[leftmargin=1.5em, itemsep=1pt]
          \item \emph{Core principles}: the model proposes, execution verifies
                (``Model Proposes, You Dispose''); parameters must be atomic
                and typed; function names must follow a Verb-Noun convention.
          \item \emph{Best practices}: use standard libraries
                (\texttt{sympy}, \texttt{numpy}) over fragile regex;
                handle edge cases explicitly; never hard-code input-specific
                values.
          \item \emph{Schema requirements}: the \texttt{name} field must
                match the top-level Python function name exactly; every
                parameter must have a \texttt{description}; the
                \texttt{required} array must list all mandatory parameters.
          \item \emph{Output format instruction}: the response must contain
                \textbf{exactly one} \texttt{```python} block and
                \textbf{exactly one} \texttt{```json} block; no other text
                is permitted.
        \end{itemize}

  \item \textbf{In-context examples.}
        $N{=}4$ question-answer pairs from the easy difficulty band of the
        current task category, formatted as:
        \begin{quote}
          \small
          \texttt{Question 1: \{q\_1\}}\textbackslash n
          \texttt{Answer: \{a\_1\}}\textbackslash n\textbackslash n
          \texttt{Question 2: \{q\_2\}}\textbackslash n
          \texttt{Answer: \{a\_2\}}\textbackslash n\textbackslash n
          \ldots
        \end{quote}

  \item \textbf{Hidden metadata.}
        A \texttt{<tool\_rl\_metadata>\ldots</tool\_rl\_metadata>} block
        containing JSON-encoded task metadata (task category, test set, ground
        truth answers).
        This block is appended by the training harness and stripped before the
        prompt is sent to the policy, so the model \emph{never} observes the
        ground-truth test answers during generation.
\end{enumerate}

\paragraph{Tool constraints.}
Every build-task instance additionally specifies two naming constraints
that are checked by the structural verifier:
the generated function must be named \texttt{solve} and must accept a single
parameter named \texttt{question}.
These constraints are stated in the instruction block and enforced by the
format reward $r^{\mathrm{fmt}}$.

\subsection{Use-Task Prompt}
\label{app:use_prompt}

The use task presents the model with a tool (in JSON schema form) and asks
it to invoke the tool to answer a question.
The model sees only the OpenAI-compatible schema, never the underlying Python
code, testing whether the schema is clear enough to drive correct invocation.

\paragraph{System message.}

\begin{AIboxC}{Use-Task System Message}
\small
You are a helpful assistant that can build tools and then use them.
\end{AIboxC}

\paragraph{Tool injection.}
The tool schemas available to the model are injected via the standard
\texttt{tools} parameter of the OpenAI chat-completion API.
Each use-task rollout receives up to $m{=}3$ schemas: 1 domain tool (from the
correct category) and 2 distractor tools (from randomly selected other
categories).
This forces the model to identify and invoke the correct schema rather than
calling the first available tool by default.

\paragraph{User message.}
The user message contains only the target question:

\begin{AIboxC}{Use-Task User Message (template)}
\small
Solve the following question. When you have the final answer, present it
as:\textbackslash n
\textbackslash boxed\{your answer here\}\textbackslash n\textbackslash n
\{question text\}
\end{AIboxC}

\paragraph{Multi-turn dialogue.}
The model responds with an OpenAI \texttt{tool\_calls} message specifying
which schema to invoke and with what arguments.
The training harness executes the Python function, formats the return value
as a \texttt{role: tool} message, and appends it to the conversation.
This continues for up to $T{=}5$ turns; the final answer must be presented
inside a \texttt{\textbackslash boxed\{\}} delimiter.

\subsection{Evaluator Prompt for $r^{\mathrm{eval}}$}
\label{app:eval_prompt}

The evaluator model $\pi^{\mathrm{eval}}$ (Sec.~\ref{sec:build}) answers
each of the $K{=}16$ held-out test questions using the generated tool,
following the same use-task prompt structure described above.
Crucially:

\begin{itemize}[leftmargin=2em, itemsep=2pt]
  \item \textbf{Only the JSON schema} is shown; the evaluator never receives
        the Python implementation.
  \item \textbf{Correctness} is verified by an LLM equivalence judge
        (Section~\ref{sec:build}) that compares the evaluator's final answer
        with the ground truth; exact string matching is used as a fast-path
        fallback.
  \item \textbf{Counting rule}: a question counts as correct only if the
        evaluator's final answer was produced via a successful tool call.
        A text-only answer (without invoking the tool) does not contribute
        to $r^{\mathrm{eval}}$, preventing the evaluator from exploiting
        its own reasoning ability to bypass the tool-use objective.
  \item \textbf{LoRA synchronisation}: $\pi^{\mathrm{eval}}$ is periodically
        refreshed by copying the latest training checkpoint weights (every
        5 gradient steps), providing an improving evaluation target as the
        policy becomes a better tool user.
        See Appendix~\ref{app:verify_judge} for a detailed discussion of the
        synchronisation artefacts this can introduce.
\end{itemize}

%% file: appendix/tool_pool_design.tex

\section{Tool Pool Design}
\label{app:tool_pool}

The Tool Pool $\mathcal{P}$ is a lightweight caching mechanism that decouples
tool creation from tool use within a single training batch.
Once a build-task rollout produces a valid tool, subsequent use-task rollouts
in the same and future batches can invoke it directly without re-running the
build step.
This section documents the pool's internal structure, admission criteria,
eviction policy, distractor selection, and initialisation.

\subsection{Pool Structure}
\label{app:pool_structure}

$\mathcal{P}$ is a thread-safe dictionary keyed by \emph{task category}:

\[
  \mathcal{P} : \text{category} \;\to\; \bigl[\,\{
    \texttt{python\_code},\;
    \texttt{openai\_tools},\;
    \texttt{quality}\}\,\bigr]
\]

Each entry stores the raw Python function string, the list of
OpenAI-compatible tool schema dicts, and the evaluation quality score
$r^{\mathrm{eval}}$ at the time of admission.
Access is protected by a single re-entrant lock, allowing concurrent rollout
workers to read and write without data races.

The task category of each rollout is parsed from the dataset row identifier
(e.g.\ the key \texttt{"bitwise\_arithmetic-train-build-42"} maps to category
\texttt{"bitwise\_arithmetic"}).

\subsection{Admission Criterion}
\label{app:pool_admission}

A tool generated during a build-task rollout is admitted to the pool if and
only if $r^{\mathrm{eval}} > 0$, i.e.\ the tool answers at least one of the
$K{=}16$ held-out test questions correctly.
Tools with $r^{\mathrm{eval}} = 0$ (including format failures and execution
errors) are discarded and never cached.

\subsection{Capacity Cap and Eviction Policy}
\label{app:pool_eviction}

Each category bucket is capped at $C{=}20$ entries.
When a new tool is admitted and the bucket is already full, the entry with the
\textbf{lowest} $r^{\mathrm{eval}}$ score is evicted to make room.
If multiple entries share the same minimum score, the \textbf{oldest} entry
(i.e.\ the one inserted earliest) is chosen as the tiebreaker, implementing
a ``quality-first, recency-as-tiebreaker'' policy.

This eviction policy has a natural curriculum effect: as training advances
and the policy writes better tools, the pool's minimum quality threshold rises
organically.
Use-task rollouts in later training steps therefore face a higher-quality
and more competitive pool than in the early steps, providing an implicit
difficulty curriculum without any explicit scheduling.

\subsection{Retrieval and Distractor Selection}
\label{app:pool_retrieval}

When a use-task rollout arrives for category $k$:

\begin{enumerate}[leftmargin=2em, itemsep=2pt]
  \item \textbf{Domain tools.}
        Up to $m_d{=}1$ tool is retrieved from $\mathcal{P}[k]$,
        taken from the \emph{most recently admitted} entries (last inserted).
        If $\mathcal{P}[k]$ is empty, the use-task rollout performs a fresh
        build pass first.

  \item \textbf{Distractor tools.}
        Up to $m_{\mathrm{dist}}{=}2$ tools are drawn from categories
        $k' \neq k$ with non-empty buckets.
        One tool is sampled uniformly at random from each eligible category,
        and the selected categories are shuffled before injection, maximising
        diversity across rollouts.

  \item \textbf{Prompt injection.}
        The domain tool and distractor tools are concatenated into a single
        \texttt{tools} list and passed to the OpenAI API, exactly as if the
        model had built all three itself.
        The model must identify the correct tool by schema inspection and
        invoke it with the right arguments.
\end{enumerate}

\noindent The combination of 1 domain tool and 2 distractors means the model
cannot succeed by calling the first tool at random: it must parse the schemas,
identify which function is relevant to the question, and construct a valid
argument dict.

\subsection{Initialisation}
\label{app:pool_init}

The pool starts \textbf{empty} at the beginning of training.
In the first training steps, all use-task rollouts that require a category
not yet covered by the pool must first perform a build pass.
The 1:1 ratio of build and use tasks in each batch
(Sec.~\ref{sec:build}) ensures that build tasks fire frequently enough in
the early steps to populate the pool quickly.

Once a category bucket crosses the threshold of at least one valid tool,
subsequent use-task rollouts in that category switch to the pool-retrieval
path and skip the build step, saving approximately one LLM forward pass per
affected rollout.

%% file: appendix/difficulty_split_table.tex

\section{Full Easy-to-Hard Difficulty Split for All 13 Training Tasks}
\label{app:difficulty_split}

Section~\ref{sec:easy_hard} describes the easy-to-hard training protocol and
gives a detailed example for the \texttt{cryptarithm} task.
This appendix provides the complete mapping for all 13 training categories:
the induction difficulty band (from which the $N{=}4$ in-context examples
are drawn) and the evaluation difficulty band (against which $r^{\mathrm{eval}}$
is computed).

The difficulty bands are defined per-task by progressively harder instantiation
parameters in Reasoning-Gym's procedural generator.
Band labels follow the generator's internal scale: higher numbers correspond to
harder instances.
For every task, the induction context uses the \emph{easiest} band (Band~2)
and the evaluation set uses the \emph{hardest} available band.
This separation ensures that a tool which merely memorises the induction
examples scores zero on the evaluation set; only a tool that encodes the
underlying algorithm generalises across the gap.

\begin{table}[h]
  \centering
  \caption{%
    Induction context and evaluation difficulty bands for all 13 SMITH training
    tasks.
    \emph{Induction band} defines the difficulty of examples shown in the
    $N{=}4$ build-task context.
    \emph{Evaluation band} defines the difficulty of instances used to compute
    $r^{\mathrm{eval}}$.
    \emph{Key difficulty axis} is the parameter that grows from easy to hard.}
  \label{tab:difficulty_split}
  \small
  \setlength{\tabcolsep}{4pt}
  \begin{tabularx}{\linewidth}{llXXX}
    \toprule
    Task & Category & Induction band (easy) & Evaluation band (hard) & Key difficulty axis \\
    \midrule
    \textit{Bitwise arithmetic}
      & Arithmetic
      & Expression depth 2
      & Expression depth 4--5
      & Nesting depth of the bit-operation expression tree \\[4pt]
    \textit{Cryptarithmetic}
      & Arithmetic
      & $\leq\!8$ unique letters (easy/medium puzzles)
      & $\geq\!9$ unique letters, including 10-letter puzzles such as
        \texttt{FORTY+TEN+TEN=SIXTY}
      & Number of unique letters (exponential search space growth) \\[4pt]
    \textit{Bit counting}
      & Algorithms
      & Integer values $1$--$10^6$
      & Integer values $10^7$--$10^8$
      & Magnitude of the input integer \\[4pt]
    \textit{LCM}
      & Algorithms
      & 2 numbers, values 1--50
      & 3--4 numbers, values 100--500
      & Count and magnitude of operands \\[4pt]
    \textit{GCD}
      & Algorithms
      & 2 numbers, values 1--500
      & 3--4 numbers, values 1{,}000--5{,}000
      & Count and magnitude of operands \\[4pt]
    \textit{Base conversion}
      & Algorithms
      & Bases 2--10, values 1--500
      & Bases 2--16, values 2{,}000--5{,}000
      & Target base range and magnitude of the number \\[4pt]
    \textit{Isomorphic string}
      & Algorithms
      & String length 10--19
      & String length 31--40
      & String length (longer strings require tracking more character mappings) \\[4pt]
    \textit{Polynomial equations}
      & Algebra
      & 2--3 terms, degree 1--2, coefficients 1--10
      & 5--6 terms, degree 3--5, coefficients 1--50
      & Polynomial degree and number of terms \\[4pt]
    \textit{Polynomial multiplication}
      & Algebra
      & 2--3 terms per polynomial, degree 1--2, 2 polynomials, coefficients 1--5
      & 5--6 terms, degree 3--5, 2--3 polynomials, coefficients 1--12
      & Degree, term count, and number of polynomials to multiply \\[4pt]
    \textit{Countdown}
      & Games
      & 4 numbers, target 10--100
      & 6 numbers, target 10--200
      & Number of available operands and target range \\[4pt]
    \textit{Tower of Hanoi}
      & Games
      & 3 disks (7 optimal moves)
      & 5 disks (31 optimal moves)
      & Number of disks (solution length grows as $2^n - 1$) \\[4pt]
    \textit{Knights and Knaves}
      & Logic
      & 2 characters, depth 2, width 3
      & 4 characters, depth 4, width 5
      & Number of characters and depth of logical deduction tree \\[4pt]
    \textit{Caesar cipher}
      & Logic
      & 3--10 words, rotation 1--10
      & 12--20 words, rotation 15--25
      & Text length and rotation offset (larger offsets are less guessable) \\
    \bottomrule
  \end{tabularx}
\end{table}

\paragraph{Training and evaluation set sizes.}
For each task, the training set (induction band) consists of 500 instances
at difficulty Band~2.
For tasks whose easy-to-hard gap is large enough to warrant it, an additional
500 instances from Band~3 (medium) are included in training, though the
evaluation set always draws from the hardest available band.
The evaluation set used to compute $r^{\mathrm{eval}}$ at each rollout step
consists of $K{=}16$ instances sampled at inference time from the hardest
difficulty band.
Final benchmark accuracy (reported in Table~\ref{tab:rg_results}) is measured
over 70 instances per difficulty level, stratified uniformly across all bands.

\paragraph{Task-category mapping.}
The 13 training categories span five higher-level groups:
\emph{Arithmetic} (\textit{bitwise arithmetic}, \textit{cryptarithmetic}),
\emph{Algorithms} (\textit{bit counting}, \textit{LCM}, \textit{GCD},
\textit{base conversion}, \textit{isomorphic string}),
\emph{Algebra} (\textit{polynomial equations}, \textit{polynomial
multiplication}),
\emph{Games} (\textit{countdown}, \textit{Tower of Hanoi}), and
\emph{Logic} (\textit{knights and knaves}, \textit{Caesar cipher}).
This spread ensures that the RL policy is exposed to a variety of algorithm
types, preventing the easy-to-hard protocol from specialising to a single
reasoning pattern.

%% file: appendix/ood_inference_protocol.tex

\section{OOD Inference Protocol for TabMWP-Hard and GQA}
\label{app:ood_inference}

Section~\ref{sec:benchmarks} describes the two out-of-domain benchmarks but
does not detail the exact inference procedure.
This appendix documents the protocol used for TabMWP-Hard and GQA, which
mirrors the Reasoning-Gym evaluation loop as closely as possible so that
differences in performance can be attributed to domain shift rather than to
evaluation asymmetry.

\subsection{Single Build-Then-Use Loop}
\label{app:ood_loop}

At evaluation time, for every benchmark the model performs a single
\emph{build} pass, then applies the resulting tool to every test instance:

\begin{enumerate}[leftmargin=2em, itemsep=3pt]

  \item \textbf{Build pass.}
        A small set of reference \texttt{(question, answer)} pairs is sampled
        from the benchmark's own training (or validation) split and presented
        to the model using the standard build-task prompt
        (Appendix~\ref{app:prompts}).
        The model generates a single Python function and a matching
        OpenAI-compatible JSON schema in one forward pass.

  \item \textbf{No retry.}
        If the build pass fails (syntax error, schema--code mismatch, or
        execution error), the tool is discarded and every test instance
        in that run receives a score of zero.
        No additional build attempts are made.

  \item \textbf{Use pass.}
        Each test question is presented to the model together with the generated
        tool's JSON schema in standard OpenAI function-calling format.
        The model may issue up to $T{=}5$ tool calls before producing a
        final answer.
        A test question scores 1 if the model invokes the tool and the
        returned answer matches the ground truth, and 0 otherwise.

\end{enumerate}

\noindent This ``single-shot build-then-use'' structure is identical to the
Reasoning-Gym evaluation loop used for RG~(Seen) and RG~(Unseen):
the same tool that was induced from easy in-context examples is applied
without modification to all test instances.

\subsection{In-Context Examples}
\label{app:ood_incontext}

In-context examples for the build pass are drawn from each benchmark's own
training split, ensuring the model has a domain-appropriate induction context:

\paragraph{TabMWP-Hard.}
We sample $N{=}4$ \texttt{(question, table, answer)} triples from the original
TabMWP training split (before augmentation).
The table is serialised as a plain-text pipe-delimited string and prepended to
the question text, matching the format of the test instances.
The same examples are used for every evaluation run.

\paragraph{GQA.}
We sample $N{=}10$ \texttt{(question, answer)} pairs from the GQA training
split.
As described in Appendix~\ref{app:gqa_setup}, no image is shown in the build
context; the model instead sees only the question strings and must infer how to
compose the visual primitives to answer them.
A larger in-context set ($N{=}10$ vs.\ $N{=}4$) is used for GQA because the
diversity of visual question types is higher and more examples are needed for
the model to infer a general strategy.

\subsection{Protocol Differences from the RG Training Loop}
\label{app:ood_diff}

Three differences distinguish the OOD evaluation protocol from the
training-time Reasoning-Gym loop:

\begin{enumerate}[leftmargin=2em, itemsep=3pt]

  \item \textbf{Easy-to-hard gap is absent.}
        For Reasoning-Gym, the induction context is drawn from easy curriculum
        bands while the evaluation set is drawn from hard bands
        (Sec.~\ref{sec:easy_hard}).
        For OOD benchmarks there is no such difficulty stratification;
        in-context examples and test instances are drawn from the same
        distribution.

  \item \textbf{Tool Pool is not used.}
        At evaluation time, no pre-built tools from the training Tool Pool are
        injected.
        The model always performs a fresh build from the provided in-context
        examples.

  \item \textbf{Evaluation reward ($r^{\mathrm{eval}}$) is not computed.}
        $r^{\mathrm{eval}}$ is a training-time signal used to grade tools
        against a held-out set of RG instances.
        At OOD evaluation time, the only signal is final test-split accuracy.

\end{enumerate}

\noindent Together, these differences mean that OOD performance reflects the
model's ability to generalise its \emph{tool-writing strategy} to new domains,
not its ability to exploit training-distribution shortcuts.

%% file: appendix/gqa_visual_tool_setup.tex

\section{GQA Visual Tool Setup}
\label{app:gqa_setup}

Because SMITH is trained exclusively on text-based Reasoning-Gym tasks, the model has
no direct perception ability at evaluation time.
To enable it to answer GQA questions nonetheless, we provide a fixed set of
three \emph{visual primitive} functions that are pre-implemented and served
via a local API.
The model's job during the build task is to compose these primitives into a
reusable tool; it never has to implement low-level vision code itself.

\subsection{Visual Primitive Functions}
\label{app:gqa_primitives}

Three primitives are available to the model:

\begin{enumerate}[leftmargin=2em, itemsep=4pt]

  \item \textbf{\texttt{locate\_objects(image\_b64, object\_name)}}:
        Uses OWL-ViT~\cite{minderer2022simple}
        (\texttt{owlvit-base-patch16}) to perform open-vocabulary
        object detection.
        Returns a list of bounding boxes in
        $[\,x_1, y_1, x_2, y_2\,]$ format for all detected instances of
        \texttt{object\_name} within the base-64-encoded JPEG image
        \texttt{image\_b64}.

  \item \textbf{\texttt{visual\_qa(image\_b64, question)}}:
        Uses BLIP-VQA~\cite{li2022blip}
        (\texttt{Salesforce/blip-vqa-base}) to answer a free-form natural
        language \texttt{question} about the image.
        Returns a short free-text answer string.

  \item \textbf{\texttt{crop\_region(image\_b64, boxes)}}:
        A pure-Python utility (no neural model) that crops the image to the
        first bounding box in \texttt{boxes}, with a $1.5\times$ padding
        margin, and returns the cropped region as a new base-64 JPEG string.

\end{enumerate}

\noindent All three primitives are served by a FastAPI server at
\texttt{localhost:8000} (configurable via the \texttt{\$GQA\_SERVER\_URL}
environment variable).

\subsection{Image Encoding}
\label{app:gqa_image}

GQA test images are loaded from the HuggingFace dataset
\texttt{[anonymous]/gqa-testdev-balanced}.
Before each tool invocation, the PIL image for the current question is
encoded as a base-64 JPEG string and bound to the Python global variable
\texttt{IMAGE}.
The generated tool code always reads from this global; it does not accept an
image path or URL argument.
This design keeps the tool interface simple
(\texttt{question: str $\to$ str}) while still providing image access:

\begin{minipage}{\columnwidth}
\begin{mdframed}
\begin{verbatim}
# Tool interface (fixed across all GQA tools)
def solve(question: str) -> str:
    # IMAGE is pre-loaded as a global b64-encoded JPEG
    objects = locate_objects(IMAGE, ...)
    answer  = visual_qa(IMAGE, question)
    return answer
\end{verbatim}
\end{mdframed}
\end{minipage}

\subsection{Build-Task Prompt for GQA}
\label{app:gqa_prompt}

Before the standard tool-maker instruction block, the build-task prompt is
prepended with a \emph{primitives context} block that (a) declares the three
available functions with their full signatures and docstrings, and (b) instructs
the model that it must compose these primitives rather than re-implementing
vision logic.
The key constraint injected into the prompt is:

\begin{quote}
\small
\textit{``Your generated tool must accept only \texttt{question: str},
compose the primitives above as needed, and return the answer as a plain
string. Do not re-implement \texttt{locate\_objects}, \texttt{visual\_qa},
or \texttt{crop\_region}; call them directly.''}
\end{quote}

\noindent The $N{=}10$ in-context examples shown to the model during the build
task are sampled uniformly at random from the GQA training split.
Each example is a \texttt{(question, answer)} pair; no image is shown in the
build-task prompt itself, so the model must reason about image content solely
through the lens of the available primitives.

\subsection{Tool Selection and Evaluation}
\label{app:gqa_selection}

Because visual tool generation is noisier than text-only tool generation
(the model has no direct visual feedback during tool writing), we generate
$M{=}10$ candidate tool proposals and select the best-performing one:

\begin{enumerate}[leftmargin=2em, itemsep=2pt]
  \item Sample $N{=}10$ reference \texttt{(question, answer)} pairs from the
        GQA training split as the in-context build context.
  \item Generate $M{=}10$ candidate Python tools using the build-task prompt.
  \item Evaluate each candidate on a disjoint validation set of 100 questions
        drawn from the GQA training split; record per-candidate validation
        accuracy.
  \item Select the candidate with the highest validation accuracy as the
        final tool.
  \item Apply the selected tool to the full GQA test split; report
        test-split accuracy.
\end{enumerate}

A failed build (syntax error or schema mismatch) receives validation accuracy
zero and is never selected unless all $M$ candidates fail, in which case a
fallback empty-response is returned and every test question scores zero.

\subsection{Answer Verification}
\label{app:gqa_verify}

GQA answers are short free-form strings (e.g.\ ``yes'', ``blue'', ``3'').
We use exact string match after lowercasing and whitespace stripping.
No LLM equivalence judge is applied; the simplicity of GQA answers makes
rule-based exact match sufficient and avoids judge overhead at the scale of
2{,}516 test questions.

%% file: appendix/retool_evaluation.tex

\section{ReTool Evaluation Setup}
\label{app:retool_setup}

ReTool~\cite{feng2025retool} is included as a distillation baseline to
isolate the contribution of the schema-grounded tool representation:
both ReTool and SMITH use execution feedback, but only SMITH produces
reusable callable schemas.
This appendix documents the specific checkpoint, training configuration, and
evaluation protocol used for our ReTool results.

\subsection{Checkpoint and Training Configuration}
\label{app:retool_checkpoint}

We fine-tune \textsc{Qwen3-4B-Instruct} on the official ReTool-SFT dataset
(\texttt{JoeYing/ReTool-SFT} on HuggingFace), which contains code-execution
trajectories distilled from \textsc{Qwen-32B}.
Training uses QLoRA with the following hyperparameters:

\begin{table}[h]
  \centering
  \caption{ReTool fine-tuning hyperparameters.}
  \label{tab:retool_config}
  \setlength{\tabcolsep}{8pt}
  \begin{tabular}{ll}
    \toprule
    Hyperparameter & Value \\
    \midrule
    Base model            & \textsc{Qwen3-4B-Instruct} \\
    Training dataset      & \texttt{JoeYing/ReTool-SFT} \\
    Adapter               & QLoRA (4-bit NF4, double quantisation) \\
    LoRA rank $r$         & 64 \\
    LoRA $\alpha$         & 128 \\
    LoRA dropout          & 0.05 \\
    LoRA target modules   & all linear layers \\
    Epochs                & 6 \\
    Sequence length       & 5{,}120 tokens \\
    Gradient accumulation & 8 \\
    Micro-batch size      & 2 \\
    Optimizer             & \texttt{paged\_adamw\_32bit} \\
    LR schedule           & Cosine \\
    Learning rate         & $4 \times 10^{-5}$ \\
    Warmup fraction       & 0.1 \\
    \bottomrule
  \end{tabular}
\end{table}

\noindent This matches the LoRA configuration used for SMITH
($r{=}64$, $\alpha{=}128$), so that any performance difference between
ReTool and SMITH is attributable to the training objective rather than the
adapter capacity.

\subsection{Evaluation Protocol on Reasoning-Gym and OOD Benchmarks}
\label{app:retool_eval_protocol}

\paragraph{Code execution.}
ReTool operates in a multi-turn loop: the model generates code in
\texttt{\seqsplit{<code>```python\ldots\\```</code>}} blocks, a sandbox executes each
block and returns output wrapped in \texttt{<interpreter>\seqsplit{ output</interpreter>}}
tags, and the model continues reasoning from the execution trace.
We allow up to $T_{\mathrm{rt}}{=}10$ code-execution rounds per question.
A question terminates when the model emits an
\texttt{<answer>\textbackslash boxed\{...\}</answer>} block, or when the
turn budget is exhausted.
The final answer is extracted from the \texttt{\textbackslash boxed\{\}}
delimiters inside the \texttt{<answer>} tag.

\paragraph{Reasoning-Gym and TabMWP-Hard.}
Each test question is presented directly to the ReTool model with its standard
system prompt; no tool schema or in-context examples are provided.
Because ReTool generates per-question code rather than a reusable schema, there
is no build pass or tool pool: every question triggers a fresh code-generation
episode.

\paragraph{GQA adaptation.}
Answering GQA questions requires access to visual primitives.
We adapt ReTool for GQA by (a) augmenting the system prompt with descriptions
of the three visual primitive functions (\texttt{locate\_objects},
\texttt{visual\_qa}, \texttt{crop\_region}; see Appendix~\ref{app:gqa_setup})
and (b) prepending the primitive implementation code to every sandbox execution
block so that \texttt{import}-free calls succeed.
With these additions, ReTool can incorporate vision into its reasoning chains
via direct function calls, on equal footing with SMITH's composed tool.

\subsection{Key Distinction from SMITH}
\label{app:retool_distinction}

ReTool and SMITH both use execution feedback, but differ in two structural
respects that are the central comparison axis of this work:

\begin{enumerate}[leftmargin=2em, itemsep=3pt]
  \item \textbf{Reusability.}
        ReTool generates ephemeral code per question; the code is discarded
        after each turn.
SMITH generates a callable JSON schema, a structured interface that
can be stored in the Tool Pool and reused across many questions without
re-running the build step.

  \item \textbf{Training signal.}
        ReTool is trained by behavioural cloning from a 32B oracle.
        SMITH is trained end-to-end from a verifiable reward, with no strong
        oracle required at training time.
\end{enumerate}

\noindent ReTool achieves the highest in-distribution accuracy on RG~(Seen)
($92.0$ vs.\ our $86.6$), reflecting the advantage of a 32B distillation
oracle on seen task families.
SMITH's advantage emerges on held-out transfer (RG Unseen, GQA, TabMWP-Hard),
where the reusable schema and RL-trained generalisation provide a structural
edge over per-question code generation.

%% file: appendix/latm_distillation.tex

\section{LATM (distill GPT-4.1) Baseline Setup}
\label{app:latm_distill}

The \textbf{LATM (4B distill GPT-4.1)} baseline tests whether behavioural
cloning from a strong frozen oracle can match RL training over an explicit
reward signal.
The oracle is GPT-4.1, which generates tool-writing trajectories for each
of the 13 training task categories; a Qwen3-4B model is then fine-tuned
on these trajectories.
This appendix documents the data collection pipeline, the fine-tuning
configuration, and the evaluation protocol.

\subsection{Trajectory Collection}
\label{app:latm_collection}

\paragraph{Oracle model.}
Tool-writing trajectories are generated by GPT-4.1 via the OpenAI API.
For each task category, GPT-4.1 is given the same build-task instruction
block used in SMITH (Appendix~\ref{app:prompts}), together with $N{=}5$
question-answer pairs sampled from the easy difficulty band.

\paragraph{Trajectory volume.}
For each task category, we collect $M{=}8$ generation attempts per
skill-set sample, where a \emph{skill set} is a particular draw of $N$ in-context
examples.
Both tool-making trajectories (oracle response = Python code + JSON schema)
and tool-use rollouts (multi-turn conversations where the tool is invoked
successfully to produce a correct answer) are retained.

\paragraph{Tool-use rollout collection.}
After tool generation, a frozen \textsc{Qwen3-4B-Instruct} model performs
up to 10 tool-call rounds for each training question using the oracle-generated
schema.
Multi-turn conversations are retained only when (a) the tool call executed
without error and (b) the produced answer matched the ground truth.
Up to 1{,}000 such conversations are collected per task category to balance
dataset size across categories.

\paragraph{Distractor augmentation.}
To teach the student model to select the correct tool among multiple candidates,
each retained tool-use conversation is augmented with distractor schemas
at two rates: 50\% of examples receive one distractor (from a randomly
chosen different category) and 90\% receive a second distractor.
Augmentation applies the same pool-retrieval logic as SMITH
(Appendix~\ref{app:tool_pool}), ensuring comparability.

The final dataset contains tool-making conversations and distractor-augmented
tool-use conversations for all 13 training categories.

\subsection{Fine-Tuning Configuration}
\label{app:latm_training}

We fine-tune \textsc{Qwen3-4B-Instruct} on the collected SFT dataset using
full-parameter fine-tuning (no LoRA adapter).
The training hyperparameters are:

\begin{table}[h]
  \centering
  \caption{LATM (distill GPT-4.1) fine-tuning hyperparameters.}
  \label{tab:latm_config}
  \setlength{\tabcolsep}{8pt}
  \begin{tabular}{ll}
    \toprule
    Hyperparameter & Value \\
    \midrule
    Base model            & \textsc{Qwen3-4B-Instruct} \\
    Adapter               & None (full fine-tuning) \\
    Epochs                & 4 \\
    Sequence length       & 16{,}000 tokens \\
    Gradient accumulation & 18 \\
    Micro-batch size      & 2 \\
    Optimizer             & \texttt{adamw\_torch} \\
    LR schedule           & Cosine \\
    Learning rate         & $4 \times 10^{-5}$ \\
    Warmup fraction       & 0.1 \\
    Training format       & Chat template; train on assistant turns only \\
    \bottomrule
  \end{tabular}
\end{table}

\noindent Fine-tuning uses full-parameter updates (not LoRA), in contrast to
SMITH's LoRA configuration ($r{=}64$, $\alpha{=}128$).
This means the LATM distilled model modifies all model weights and may have a
larger effective capacity for memorising the oracle's style.
The comparison therefore tests \emph{training signal quality} (RL reward vs.\
behavioural cloning) rather than model capacity.

\subsection{Evaluation Protocol}
\label{app:latm_eval}

At evaluation time, the fine-tuned LATM model is evaluated using the standard
build-then-use inference protocol
(Appendix~\ref{app:ood_inference}).
No oracle model is involved at test time; the student model generates tools
autonomously from the $N{=}4$ in-context examples.
Results are reported across RG~(Seen), RG~(Unseen), TabMWP-Hard, and GQA
in Table~\ref{tab:rg_results} and Table~\ref{tab:ood_transposed}.

\subsection{Comparison to SMITH}
\label{app:latm_vs_smith}

The key structural difference between LATM (distill GPT-4.1) and SMITH is the
source of the training signal:

\begin{itemize}[leftmargin=2em, itemsep=3pt]
  \item \textbf{LATM (distill GPT-4.1)} trains by imitating GPT-4.1 outputs.
        The student learns to reproduce what GPT-4.1 writes, regardless of
        whether those tools actually execute correctly on held-out instances.

  \item \textbf{SMITH} trains from a verifiable execution reward.
        Every gradient update is conditioned on whether the generated tool
        answered the held-out test questions correctly; no oracle demonstrations
        are required.
\end{itemize}

\noindent LATM (distill GPT-4.1) achieves strong in-distribution accuracy
($83.3$ on RG Seen) because GPT-4.1 generates high-quality tools for the
training task categories.
SMITH's advantage emerges on transfer benchmarks (RG Unseen $79.8$ vs.\
$66.4$; GQA $42.6$ vs.\ $\mathbf{56.0}$\textsuperscript{\dag}), where the
execution reward drives the policy toward tools that generalise
beyond the oracle's demonstrated style.

\noindent\footnotesize\textsuperscript{\dag}The GPT-4.1 distillation gap on GQA
reflects that GPT-4.1 has native visual understanding and generates visual
primitives the 4B student can imitate; SMITH reaches its GQA score entirely
through the self-supervised tool-creation loop without visual oracle data.

%% file: appendix/baseline_agentic_comparison.tex
\section{Agentic Loop Comparison: KTCE, CRAFT, TroVE, and SMITH}
\label{app:agentic_comparison}

This appendix provides a detailed, implementation-level comparison of the agentic
designs of the three baseline systems evaluated in Section~\ref{sec:baselines}
(KTCE~\cite{ma2025automated}, CRAFT~\cite{yuancraft}, and
TroVE~\cite{wang2024trove}) alongside our own SMITH framework.
The central axis of variation is \emph{when} tool creation happens relative to
test-time inference.

\subsection{Positioning Overview}

Table~\ref{tab:tool_creation_timing} situates each system along the tool-creation
timeline before discussing the per-system loops in detail.

\begin{table}[h]
  \centering
  \caption{Tool-creation timing, model modification, and cross-problem reuse for
    each system.}
  \label{tab:tool_creation_timing}
  \setlength{\tabcolsep}{5pt}
  \begin{tabular}{lccc}
    \toprule
    \textbf{System} & \textbf{Tool creation timing} & \textbf{Weights changed?} & \textbf{Tools survive across problems?} \\
    \midrule
    KTCE   & Fully offline (before any test problem) & No  & Yes (fixed toolset) \\
    CRAFT  & Fully offline (before any test problem) & No  & Yes (fixed library) \\
    TroVE  & Online / streaming (during test inference) & No  & Yes (library grows sequentially) \\
    SMITH  & Trained offline via RL                  & \textbf{Yes} & Yes (quality-filtered pool) \\
    \bottomrule
  \end{tabular}
\end{table}

\subsection{KTCE: Offline Evolutionary Creation with Online Retrieval-then-Solve}
\label{app:ktce_loop}

KTCE separates tool creation from inference with a hard boundary.
All tool work happens before any test problem is seen.

\paragraph{Offline phase.}
Training problems are grouped by mathematical subfield using BGE-M3 embeddings
and $k$-means clustering.
For each subfield cluster, an LLM generates candidate Python functions; semantically
near-duplicate candidates are collapsed via agglomerative clustering
(similarity $\geq 0.80$), and one verified tool per cluster is selected by
majority-vote execution.
This initial toolset then enters a 5-iteration evolutionary loop:

\begin{enumerate}[leftmargin=2em, itemsep=2pt]
  \item \textbf{Evaluate}: run all training problems in the subfield through
        the current toolset; record per-tool usage frequency (\texttt{Freq})
        and tool success rate (\texttt{TSR}).
        Compute a composite loss $\alpha \sum Q_{\text{tool}} + \beta Q_{\text{set}}
        + \gamma \max(0, n - k)$.
  \item \textbf{Delete}: LLM decides which tools (up to 3--5) to remove based
        on low frequency and success rate.
  \item \textbf{Modify}: for each tool with $\text{TSR}/\text{Freq} \leq 0.90$,
        generate an evolved version using failure examples as context;
        validate by execution before substituting.
  \item \textbf{Add}: LLM proposes new tools for problems currently uncovered;
        validated by execution before insertion.
  \item \textbf{Rollback}: if loss increases, revert to the previous iteration
        and pass failure context to the next modify/add step.
\end{enumerate}

Each tool record carries a natural-language \texttt{experience\_pool}: usage
examples accumulated during the evaluate step.
The final toolset is organised as a two-level map:
$\text{Field} \to \text{Subfield} \to [\text{tool}]$.

\paragraph{Online phase (per test problem).}

\begin{enumerate}[leftmargin=2em, itemsep=2pt]
  \item Retrieve the subfield from per-problem metadata.
  \item \textbf{LLM call 1}: model reads a numbered list of subfield tools
        and selects which to use; output parsed for tool indices.
  \item \textbf{LLM call 2}: model generates Python code that calls the
        selected tools, augmented by BGE-M3-ranked few-shot examples from the
        experience pool.
        Code is executed; output (\texttt{hint}) is captured.
  \item \textbf{LLM call 3}: chain-of-thought extraction of the final answer
        from the hint.
\end{enumerate}

\textbf{LLM calls at test time: 3} (retrieval select + code-gen + CoT extract).

\subsection{CRAFT: Offline Diversity-Sampled Creation with Multi-View Retrieval}
\label{app:craft_loop}

CRAFT constructs its tool library offline using GPT-4 for both creation and
abstraction steps, then switches to GPT-3.5-turbo for inference.

\paragraph{Offline phase.}
Training problems are sampled in diversity-maximising epochs: epoch~0 draws
200 random problems; each subsequent epoch ranks remaining problems by their
minimum SimCSE cosine similarity to already-sampled problems and takes the 100
most dissimilar.
For each sampled problem:

\begin{enumerate}[leftmargin=2em, itemsep=2pt]
  \item \textbf{LLM call 1 (GPT-4)}: generate a specific Python solution for
        this problem.
        Execute and grade; discard if incorrect.
  \item \textbf{LLM call 2 (GPT-4)}: abstract the specific solution into a
        general parametric function with a docstring.
        Execute the abstract tool; discard if non-executable.
\end{enumerate}

Surviving tools are deduplicated by function-name similarity.
Each tool is then encoded into four SimCSE vectors:
(i) function name, (ii) one-line explanation, (iii) full docstring,
(iv) the original problem that generated the tool.
These are stored as \texttt{vector\_library.pt} for retrieval.

\paragraph{Online phase (per test problem).}

\begin{enumerate}[leftmargin=2em, itemsep=2pt]
  \item Encode the test question with SimCSE; score against all four vector
        views in the library; retrieve top-$k$ tools ($k=3 \to 2 \to 1$ if
        context overflows).
  \item \textbf{LLM call 1}: question + retrieved tool code (pasted inline)
        $\to$ generate a new specific solution that may call or extend the
        retrieved functions.
  \item \textbf{LLM call 2}: question + new solution $\to$ generate the
        API call expression.
  \item Execute retrieved tools + new code + API call; grade against ground
        truth.
\end{enumerate}

\textbf{LLM calls at test time: 2} (code-gen + API call).

\subsection{TroVE: Online Streaming Creation with Frequency-Based Forgetting}
\label{app:trove_loop}

TroVE has no offline pre-processing phase.
The library starts from a minimal default toolbox (standard imports) and grows
as test problems are processed in order.

\paragraph{Online phase (per test problem, processed sequentially).}

Three modes are run in parallel for each problem:

\begin{enumerate}[leftmargin=2em, itemsep=2pt]
  \item \textbf{Import mode}: the full \emph{current} library is injected
        into the prompt; the LLM generates a solution using existing functions.
  \item \textbf{Create mode}: only the default (initial) library is shown;
        the LLM generates a solution \emph{and} defines new helper functions it
        believes can be reused.
  \item \textbf{Skip mode}: no library is shown; the LLM generates a plain
        Python solution.
\end{enumerate}

Each mode's response is executed and graded.
The best response (by correctness, then executability) is selected:

\begin{itemize}[leftmargin=2em, itemsep=2pt]
  \item If \textbf{import} wins: increment the usage-frequency counter for each
        called function.
  \item If \textbf{create} wins and execution succeeded: add the new functions
        to the library.
\end{itemize}

\paragraph{Periodic forgetting.}
Every 500 problems, tools whose frequency falls below
$\log_{20}(n)$ (where $n$ is the number of problems processed so far) are
pruned from the library.
Problems whose winning tool was pruned are queued for re-generation using only
the import and skip modes after all problems have been processed.

\textbf{LLM calls at test time: 3 per problem} (one per mode, run in parallel).
Later problems benefit from tools created for earlier ones; the result therefore
depends on problem order.

\subsection{SMITH: RL-Trained Tool Creation Coupled with Tool Use}
\label{app:smith_loop}

SMITH does not prompt a frozen model to write tools at test time.
Instead, it trains a 4B model via RL to be a capable tool creator, jointly
optimising tool creation and tool use inside a single learning objective.

\paragraph{Training loop.}
Training alternates between two task types on 13 procedural task categories
from Reasoning-Gym:

\begin{itemize}[leftmargin=2em, itemsep=2pt]
  \item \textbf{Build tasks}: given a task description and a small set of
        in-context examples, generate a Python function and a matching
        OpenAI-compatible JSON schema in a single forward pass.
        Three independent reward signals are computed: execution accuracy on
        held-out questions ($r_{\text{eval}}$), LLM-as-judge code quality
        ($r_{\text{judge}}$, see Appendix~\ref{app:judge_prompt}), and format
        consistency of the JSON schema ($r_{\text{format}}$).
        Correctness reward is granted \emph{only} when the final answer is
        produced through a successful tool invocation, removing any incentive
        to substitute text-only reasoning.
  \item \textbf{Use tasks}: given a tool from the shared pool, invoke it
        correctly to answer a held-out question; reward is rule-based answer
        matching.
\end{itemize}

The shared pool admits tools only after they pass execution evaluation.
As the pool fills, weaker tools are evicted, creating an implicit curriculum
that exposes the policy to progressively stronger competition.

\paragraph{Online phase (per test problem).}

\begin{enumerate}[leftmargin=2em, itemsep=2pt]
  \item \textbf{Build}: 1 LLM pass generates the Python function and JSON
        schema.
  \item \textbf{Use}: LLM invokes the tool via the schema; execution
        returns the result; LLM synthesises the final answer.
\end{enumerate}

\textbf{LLM calls at test time: ${\sim}$3} (build + invoke + answer synthesis).
Unlike all baselines, the model has been \emph{trained} to write the kind of
tool it can also reliably invoke.

\subsection{Full Comparison Table}
\label{app:comparison_table}

Table~\ref{tab:full_comparison} summarises all dimensions across the four
systems.

\begingroup
\small
\setlength{\tabcolsep}{4pt}
\begin{longtable}{>{\raggedright\arraybackslash}p{3.3cm}
    >{\raggedright\arraybackslash}p{\dimexpr(5.5in-3.3cm-10\tabcolsep)/4\relax}
    >{\raggedright\arraybackslash}p{\dimexpr(5.5in-3.3cm-10\tabcolsep)/4\relax}
    >{\raggedright\arraybackslash}p{\dimexpr(5.5in-3.3cm-10\tabcolsep)/4\relax}
    >{\raggedright\arraybackslash}p{\dimexpr(5.5in-3.3cm-10\tabcolsep)/4\relax}}
  \caption{Implementation-level comparison of agentic tool-creation designs.}
  \label{tab:full_comparison}\\
  \toprule
  \textbf{Dimension} & \textbf{KTCE} & \textbf{CRAFT} & \textbf{TroVE} & \textbf{SMITH} \\
  \midrule
  \endfirsthead
  \multicolumn{5}{c}{\tablename~\thetable\ (continued)}\\
  \toprule
  \textbf{Dimension} & \textbf{KTCE} & \textbf{CRAFT} & \textbf{TroVE} & \textbf{SMITH} \\
  \midrule
  \endhead
  \midrule
  \endfoot
  \bottomrule
  \endlastfoot

  \textbf{Tool creation trigger}
    & Per knowledge-subfield cluster of training data
    & Per training problem (diversity-sampled across epochs)
    & Per test problem that yields a novel, executable function
    & RL training rollout on build tasks \\[4pt]

  \textbf{Tool structure}
    & Python function + name / docstring / \texttt{experience\_pool}
    & Python function + docstring + 4 SimCSE embedding vectors
    & Python function + docstring + frequency counter
    & Python function + OpenAI JSON schema \\[4pt]

  \textbf{Tool verification}
    & Execution + majority-vote during creation; loss-tracked during optimization
    & Execution + answer correctness at both specific and abstract stages
    & Implicit: tool enters library only if full solution executes correctly
      and wins 3-way selection
    & Execution accuracy as RL reward; format consistency as separate reward axis \\[4pt]

  \textbf{Optimization / refinement}
    & Explicit 5-iteration evolutionary loop: evaluate $\to$ compute loss $\to$
      LLM delete / modify / add $\to$ rollback if worse
    & None (one-shot: create, validate, deduplicate)
    & Implicit frequency-based forgetting: low-reuse tools pruned every 500
      examples
    & RL training is the optimization loop; gradient updates improve the
      tool-writing policy itself \\[4pt]

  \textbf{Retrieval mechanism}
    & Two-stage: (1)~subfield lookup from metadata; (2)~LLM reads numbered
      list and selects tools
    & Multi-view SimCSE similarity across 4 views (name, explanation,
      docstring, original question)
    & None; entire current library injected into the import-mode prompt,
      library size bounded by trimming
    & None at test time; model generates the required tool directly
      (trained to do so) \\[4pt]

  \textbf{Tool reuse across problems}
    & Yes (fixed toolset shared across all test problems)
    & Yes (fixed library shared across all test problems)
    & Yes (tools created for problem $i$ available for problem $i{+}1$
      onward)
    & Yes (shared execution-verified pool) \\[4pt]

  \textbf{Order dependency}
    & No (offline toolset is order-independent)
    & No (offline library is order-independent)
    & \textbf{Yes}; later problems benefit from tools created for earlier
      ones, and shuffling changes results
    & No at inference (pool is pre-built) \\[4pt]

  \textbf{Agent loop at inference}
    & Linear: subfield lookup $\to$ LLM selects tools $\to$ LLM generates
      code $\to$ execute $\to$ LLM CoT extract
    & Linear: multi-view retrieve $\to$ LLM generates code (may extend tool)
      $\to$ LLM generates API call $\to$ execute
    & 3-way parallel per problem (import / create / skip) $\to$ select best
      $\to$ conditionally update library
    & Build: 1-pass generation of function + schema; Use: invoke $\to$ execute
      $\to$ synthesise answer \\[4pt]

  \textbf{LLM calls at test time}
    & 3 (retrieval select + code-gen + CoT extract)
    & 2 (code-gen + API call)
    & 3 (one per parallel mode)
    & ${\sim}$3 (build + invoke + answer synthesis) \\[4pt]

  \textbf{LLM calls during tool creation}
    & 100s--1000s total (5 iters $\times$ $N$ problems $\times$ multiple
      calls per iter, per subfield)
    & 2 per training sample (specific solution + abstraction), both GPT-4
    & 0 (no offline phase)
    & RL training rollouts (amortised into model weights) \\[4pt]

  \textbf{Semantic embeddings}
    & BGE-M3 (clustering, dedup, few-shot retrieval)
    & SimCSE (diversity sampling + multi-view retrieval)
    & None
    & None \\[4pt]

  \textbf{Training required?}
    & No
    & No
    & No
    & \textbf{Yes} (RL fine-tuning) \\[4pt]

  \textbf{Primary model(s)}
    & GPT-3.5-turbo (retrieval, solve, evolve); BGE-M3 for embeddings
    & GPT-4 (construction); GPT-3.5-turbo (inference); SimCSE (retrieval)
    & CodeLlama-7b (default) or any OpenAI-compatible model
    & 4B RL-trained model \\

\end{longtable}
\endgroup

\subsection{Key Conceptual Distinctions}
\label{app:key_distinctions}

\paragraph{What the LLM sees at inference.}
The four systems differ fundamentally in how a tool is \emph{presented} to the
model at solve time:

\begin{itemize}[leftmargin=2em, itemsep=3pt]
  \item \textbf{KTCE} pastes tool code, docstring, and experience-pool examples
        inline into the solution prompt; the LLM writes new code that calls the
        tool functions directly.
  \item \textbf{CRAFT} pastes retrieved tool code inline; the LLM writes new
        code that may call or extend the retrieved functions.
  \item \textbf{TroVE} pastes the \emph{entire current library} of function
        definitions into the import-mode prompt; the LLM writes code that
        imports from the toolbox by name.
  \item \textbf{SMITH} presents the tool as an OpenAI function-calling schema
        (not raw code); the LLM issues a \texttt{tool\_calls} message, the
        Python function is invoked externally, and the result is returned as a
        \texttt{tool} role message before the LLM synthesises the final answer.
\end{itemize}

\paragraph{The optimization target.}
KTCE and TroVE both maintain an evolving tool library, but their optimization
strategies are orthogonal: KTCE applies an \emph{explicit} LLM-driven
delete/modify/add loop with loss-guided rollback, while TroVE applies
\emph{implicit} population pressure through frequency-based forgetting.
CRAFT applies no post-creation optimization.
SMITH's optimization is the RL training process itself; rather than refining
individual tools after the fact, gradient descent directly improves the
\emph{policy} that generates tools, making each new tool better than the last.

\paragraph{The retrieval bottleneck.}
KTCE and CRAFT require a retrieval step before any tool can be used; retrieval
quality therefore bounds solution quality.
TroVE sidesteps this by injecting the entire library into the prompt, but this
only works because frequency-based trimming keeps the library small enough to
fit in context.
SMITH eliminates retrieval entirely: the trained model writes the tool it needs
from scratch in one pass, bypassing any index or library lookup.

%% file: appendix/baseline_failure_analysis.tex
\section{Baseline Failure Analysis}
\label{app:baseline_failure}

This appendix documents implementation-level failure modes we observed when
evaluating KTCE~\cite{ma2025automated} and TroVE~\cite{wang2024trove} with
Qwen3-4B-Instruct-2507 on our benchmark suite.
These observations inform the token-cost and accuracy numbers reported in
Table~\ref{tab:rg_results} and motivate several methodological choices in
our evaluation pipeline.

\subsection{KTCE: Three Compounding Failure Modes}
\label{app:ktce_failures}

Inspecting per-question inference outputs across all 24 tasks reveals three
distinct failure modes that together explain KTCE's uneven task profile.

\paragraph{Failure Mode 1: Programmatic solver bypass.}
KTCE's inference code falls back to a hand-coded \texttt{solve} dispatcher
when no LLM is configured or when the tool retrieval step returns an empty
set.  Tasks such as \texttt{bitwise\_arithmetic}, \texttt{count\_bits},
\texttt{tower\_of\_hanoi}, and \texttt{isomorphic\_string} are solved entirely
by this dispatcher, reaching 100\% accuracy without a single LLM call.
While this inflates overall averages, it also masks the failure of the
tool-generation pipeline for those categories: the evolutionary toolset
produced for these tasks was never invoked.

Conversely, tasks whose programmatic solver is incomplete or absent drop to
near-zero: \texttt{cryptarithm} (5.7\%), \texttt{knights\_knaves} (4.8\%),
and \texttt{polynomial\_equations} (1.1\%) all fall into this category.
\texttt{GQA} reaches 0\% because the dispatcher has no implementation for
visual question answering.

\paragraph{Failure Mode 2: Stub tool generation.}
For approximately half the tasks in our suite, KTCE's offline evolutionary
loop produces a degenerate tool, specifically a function whose body is simply
\texttt{return ""}.  A representative example from the \texttt{ab} task:

\begin{verbatim}
def solve_ab(problem: str) -> str:
    return ""
\end{verbatim}

When the LLM subsequently generates code that calls \texttt{solve\_ab}, code
execution succeeds but yields an empty string, causing the grader to mark the
answer incorrect.  Accuracy for stub-tool tasks therefore depends entirely on
whether \texttt{\_extract\_final\_answer} can recover a usable answer from the
raw model response, effectively reducing KTCE to a plain chain-of-thought
baseline.  Tasks where the model can reason textually without the tool still
score well (\texttt{syllogism}: 96.7\%, \texttt{gcd}: 96.2\%), while
computation-heavy tasks that genuinely need a working tool collapse
(\texttt{ab}: 0\%, \texttt{group\_anagrams}: 0\%, \texttt{base\_conversion}: 0\%).

\paragraph{Failure Mode 3: LLM endpoint instability.}
Inspecting individual inference-output files reveals that many runs experienced
intermittent LLM failures: the API call returned an empty response, causing
\texttt{\_solve\_with\_llm} to return early with no generated code and no
token-usage record.  For example, the \texttt{ab} task has only 64 of 210
entries with a real LLM response.  This also corrupts the aggregated
\texttt{all\_results.json}, which can be silently overwritten by a subsequent
failed re-run, replacing real API token counts with tokenizer-estimated zeros.
We therefore read token usage from individual \texttt{inference\_output/*.json}
files and skip entries where \texttt{completion\_tokens} = 0 when computing
the Avg.~Tokens figure in Table~\ref{tab:rg_results}.

\paragraph{Per-task breakdown.}
Table~\ref{tab:ktce_per_task} summarises all three failure modes across the
full task suite.

\begin{table}[h]
  \centering
  \caption{Per-task KTCE diagnostic breakdown (Qwen3-4B-Instruct-2507).
    \emph{Real LLM calls} counts entries with \texttt{completion\_tokens} $>0$
    in the per-question inference output files.
    \emph{Stub tool} indicates the generated tool body unconditionally returns
    an empty string.}
  \label{tab:ktce_per_task}
  \small
  \setlength{\tabcolsep}{5pt}
  \begin{tabular}{lrrrll}
    \toprule
    Task & Acc.\ (\%) & N & Real LLM calls & Stub tool & Failure mode \\
    \midrule
    ab                     &  0.0 & 210 &  64 / 210 & Yes & FM2 + FM3 \\
    base\_conversion       &  0.0 & 210 &   0 / 210 & Yes & FM2 + FM3 \\
    bitwise\_arithmetic    & 100.0 & 280 &   0 / 280 & N/A & FM1 (solver) \\
    caesar\_cipher         & 82.4 & 210 &   0 / 210 & No & FM1 (solver) \\
    calendar\_arithmetic   & 66.7 & 198 & 198 / 198 & Yes & FM2 \\
    chinese\_theorem       & 100.0 & 100 &   0 / 100 & No & FM1 (solver) \\
    complex\_arithmetic    & 25.0 & 200 & 200 / 200 & Yes & FM2 \\
    count\_bits            & 100.0 & 210 &   0 / 210 & No & FM1 (solver) \\
    countdown              & 78.6 & 210 & 210 / 210 & Yes & FM2 \\
    cryptarithm            &  5.7 & 210 &   0 / 210 & No & FM1 (incomplete solver) \\
    gcd                    & 96.2 & 210 & 210 / 210 & Yes & FM2 \\
    gqa                    &  0.0 & 2516 &  0 / 2516 & No & FM1 (no solver) \\
    group\_anagrams        &  0.0 & 200 & 200 / 200 & Yes & FM2 \\
    gsm8k                  & 50.9 & 1319 &  0 / 1319 & No & FM1 (solver) \\
    isomorphic\_string     & 100.0 & 350 &   0 / 350 & No & FM1 (solver) \\
    knights\_knaves        &  4.8 & 210 &   0 / 210 & No & FM1 (incomplete solver) \\
    lcm                    & 18.1 & 210 & 210 / 210 & Yes & FM2 \\
    polynomial\_equations  &  1.1 & 280 &   0 / 280 & No & FM1 (incomplete solver) \\
    polynomial\_mult.      & 66.1 & 280 &   0 / 280 & No & FM1 (solver) \\
    puzzle24               & 40.1 & 382 & 382 / 382 & Yes & FM2 \\
    self\_reference        & 75.6 & 234 & 234 / 234 & Yes & FM2 \\
    simple\_equations      & 16.0 &  16 &  16 /  16 & Yes & FM2 \\
    syllogism              & 96.7 & 210 & 210 / 210 & Yes & FM2 \\
    tabmwp                 & 56.0 & 3152 &  0 / 3152 & No & FM1 (solver) \\
    tower\_of\_hanoi       & 100.0 & 210 &   0 / 210 & No & FM1 (solver) \\
    \bottomrule
  \end{tabular}
\end{table}

\subsection{TroVE: Missing Token-Usage Records}
\label{app:trove_token}

TroVE uses a two-phase pipeline: Phase~1 (validate split) grows the tool
library; Phase~2 (test split, frozen library) is used for scoring.
Token usage is stored in the results files only when the OpenAI-compatible
backend records it.
Inspecting the Phase~2 results across 20 tasks, 12 tasks have no stored
\texttt{token\_usage} whatsoever.

The \texttt{load\_trove\_token\_average} function previously fell back to
reconstructing the prompt by rendering TroVE's Mako templates with the
\emph{initial} three-function toolbox.
However, Phase~2 prompts actually include the \emph{learned} library from
Phase~1, which can be substantially larger.
Using the initial toolbox for reconstruction produces a prompt of only
${\sim}$179 characters (${\sim}$45 tokens), far smaller than the real
prompt, leading to a severely underestimated average of 691 tokens.

We correct this by computing the macro-average token count \emph{only} over
tasks that have real API-recorded usage (8 of 20 tasks), yielding
829 tokens.
This figure is itself a lower bound: the 12 excluded tasks plausibly have
larger prompts due to larger learned libraries, so the true TroVE average
likely exceeds 829 tokens per question.

%% file: appendix/tabmwp_hard_dataset.tex

\section{TabMWP-Hard: Dataset Construction and Augmentation}
\label{app:tabmwp_hard}

Standard CoT achieves $96.8\%$ on the original TabMWP benchmark
(Table~\ref{tab:rg_results}), making it an unreliable signal of tabular
reasoning ability: a model that simply identifies the named entity in a
small, clean table and reads off its value will score near-perfectly.
This appendix documents the augmentation pipeline used to construct
\textbf{TabMWP-Hard}, a harder evaluation set derived from the same
problems, and illustrates each transformation step with concrete examples
from the dataset.

\subsection{Table Type Coverage}
\label{app:tabmwp_types}

TabMWP tables are first assigned to one of seven structural types via a
deterministic rule-based classifier (priority-ordered: \textsc{Stem-and-Leaf},
\textsc{Financial Ledger}, \textsc{Two-Way}, \textsc{Function Table},
\textsc{Price Rate}, \textsc{Price List}, \textsc{Named Count}).
We select \textbf{three} types for full augmentation:
\textsc{Financial Ledger}, \textsc{Price List}, and \textsc{Two-Way}.

\textsc{Stem-and-Leaf} tables are excluded because the key space is
inherently bounded: stems are single digits 0–9, so at most
$10 - |\text{existing stems}|$ distractor rows can ever be added.
A table with five existing stems leaves room for at most five new ones,
far too few to bury the target row or challenge a model's lookup ability.
\textsc{Function Table} entries are similarly excluded because distractor
rows must extend the exact linear sequence, leaving no freedom to
generate confusingly adjacent keys.
The three selected types admit arbitrarily large distractor pools and
support both large-scale row injection and adversarial near-miss
construction.

\subsection{Augmentation Pipeline}
\label{app:tabmwp_pipeline}

Each selected entry passes through four sequential stages:

\begin{enumerate}[noitemsep, topsep=2pt]
  \item \textbf{Distractor row injection.} Up to 2,000 type-compatible
        rows are inserted at random positions, growing the table from a
        handful of rows to hundreds or thousands.
  \item \textbf{Near-miss row insertion.} An LLM (or rule-based
        fallback) generates rows whose first-column key is \emph{similar
        but distinct} from the key the question asks about, and places
        them immediately adjacent to the target row.
  \item \textbf{LLM column augmentation.} A language model proposes
        3–4 additional columns that are contextually plausible for the
        table's domain but do not help answer the question.
        For \textsc{Price List} tables, which carry no header in their
        original format, the LLM simultaneously names the existing
        columns and invents new ones, injecting a proper header row.
  \item \textbf{Two-step validity gate.} The LLM first answers the
        question from the \emph{original} table to confirm it can reach
        the correct answer, then verifies the augmented table still
        contains the necessary information.
        Entries that fail either step are discarded.
\end{enumerate}

\noindent Entries that survive all four stages are further checked for
key collisions (Section~\ref{app:tabmwp_collision}) before being written
to the final dataset.

\subsection{Worked Example A: \textsc{Two-Way} Table}
\label{app:tabmwp_ex_twoway}

\paragraph{Original table.}
The original entry is a five-row, three-column philanthropic-donation
table.  The question asks for the difference between two specific people's
donations to a specific cause.

\begin{AIboxC}{Original table: \textsc{Two-Way} (5 rows $\times$ 3 columns)}
\small
\begin{verbatim}
Person         | Animal rights | Clean water
Eve            | $4            | $15
Eli            | $12           | $5
Bridgette      | $9            | $11
Kamal          | $18           | $11
Janelle        | $13           | $13
\end{verbatim}
\textbf{Q:} How much more money did Eve donate to clean water than Eli?
\quad \textbf{A:} \$10
\end{AIboxC}

\noindent Answering requires only two lookups in a clean, five-row table,
trivial for any language model that can parse basic tabular text.

\paragraph{After distractor injection.}
2,000 rows are requested; the generator samples entity names from a
dataset-wide pool, produces per-column values within each column's
observed type and range (integer currency for this table), and shuffles
all rows into random positions.
The target rows (Eve and Eli) are buried at arbitrary offsets.

\paragraph{After near-miss insertion.}
The LLM identifies ``Eve'' and ``Eli'' as the two keys referenced by
the question and inserts similarly-named rows immediately adjacent to
each target.  Because this is a person-name table, the generated
near-miss keys are phonetically or orthographically close (e.g.\
``Ev'', ``Elliot'') with deliberately different dollar values.

\paragraph{After LLM column augmentation.}
The LLM proposes four additional columns that fit a philanthropic
context but do not encode the answer: \textit{Membership Year},
\textit{Donation Method}, \textit{Annual Giving Level}, and
\textit{Recurring Donor}.  Python snippets supplied by the LLM fill
all rows with plausible values.
The final table has \textbf{7 columns} and \textbf{308 rows}.

\begin{AIboxC}{Augmented table: \textsc{Two-Way} (308 rows $\times$ 7 columns, excerpt)}
\scriptsize
\tiny
\begin{verbatim}
Person         | Animal rights | Clean water | Membership Year | Donation Method | Annual Giving Level | Recurring
...            | ...           | ...         | ...             | ...             | ...                 | ...
Patty          | $11           | $6          | 2013            | Check           | Silver              | F
Sports         | $4            | $13         | 2018            | Bank Transfer   | Platinum            | T
Eve            | $4            | $15         | 2018            | PayPal          | Gold                | T← target
Ev             | $6            | $9          | 2019            | Bank Transfer   | Silver              | F← near-miss
Elliot         | $10           | $11         | 2021            | Credit Card     | Bronze              | F← near-miss
Trisha         | $7            | $10         | 2015            | Check           | Platinum            | T
...            | ...           | ...         | ...             | ...             | ...                 | ...
\end{verbatim}
\end{AIboxC}

\noindent A model reading this table must: (1) locate the correct row
among hundreds, (2) ignore four irrelevant columns to read the right
value, and (3) not be misled by the adjacent near-miss entries whose
names differ by only one or two characters.

\subsection{Worked Example B: \textsc{Price List} Table (Missing Header)}
\label{app:tabmwp_ex_pricelist}

\paragraph{Original table.}
\textsc{Price List} tables in the original TabMWP dataset carry
\textbf{no header row}: every line is a data row in the format
\texttt{item name | \$price}.
The following entry has four rows and asks about the combined cost of
two items.

\begin{AIboxC}{Original table: \textsc{Price List} (4 rows, \textbf{no header})}
\small
\begin{verbatim}
orange cone shell  | $0.05
spiral snail shell | $0.03
purple clam shell  | $0.03
scallop shell      | $0.08
\end{verbatim}
\textbf{Q:} Cassie has \$0.18.  How much money will Cassie have left
if she buys a spiral snail shell and a scallop shell? \quad \textbf{A:} \$0.07
\end{AIboxC}

\paragraph{After distractor injection.}
New item–price rows are generated from a dataset-wide item pool,
matched to the existing whole-cent price format and \$0.03–\$0.08
range.
Items are paired to form compound names (e.g.\ \texttt{"bag of peanuts
digital camera"}) so that distractor keys can never accidentally match
a single-word original key.

\paragraph{After near-miss insertion.}
The LLM generates rows whose names are \emph{semantically related but
distinct} from the two answer items (``spiral snail shell'' and ``scallop
shell'').
Because size or variant suffixes (e.g.\ ``spiral snail shell (small)'')
are explicitly forbidden in the prompt, the LLM instead invents
natural-language variants: \textit{spiral whelk shell},
\textit{striped snail shell}, \textit{scallop valve},
\textit{giant scallop shell}.
These are inserted near the target rows so that a model skimming for
``shell'' will encounter multiple plausible but incorrect candidates.

\paragraph{After LLM column augmentation (with header injection).}
Because the original table has no header, the LLM is asked to perform
two tasks simultaneously: (a) name the two existing columns, and (b)
propose 3–4 new columns.
It returns names for the existing columns (\textit{Shell Type},
\textit{Unit Price}) plus new columns (\textit{Inventory Count},
\textit{Origin Region}, \textit{Shell Grade}, \textit{Supplier ID}).
A header row is constructed and prepended to the table.
The final table has \textbf{6 columns} and \textbf{316 rows}.

\begin{AIboxC}{Augmented table: \textsc{Price List} (316 rows $\times$ 6 columns, excerpt)}
\scriptsize
\begin{verbatim}
Shell Type           | Unit Price | Inventory Count | Region | Grade | Supplier ID
...                  | ...        | ...             | ...    | ...    | ...
bag of peanuts ...   | $0.06      | 126             | Arctic | C      | SUP-1083
scallop shell        | $0.08      | 405             | Atlantic| C     | SUP-1084  ← target
night's stay at ...  | $0.05      | 158             | Pacific | B     | SUP-1085
...                  | ...        | ...             | ...     | ...   | ...
spiral snail shell   | $0.03      | 400             | Arctic  | C     | SUP-1232  ← target
spiral whelk shell   | $0.06      | 10              | Atlantic| B     | SUP-1240  ← near-miss
striped snail shell  | $0.04      | 316             | Arctic  | C     | SUP-1241  ← near-miss
scallop valve        | $0.07      | 175             | Indian  | C     | SUP-1242  ← near-miss
giant scallop shell  | $0.12      | 260             | Pacific | C     | SUP-1243  ← near-miss
...                  | ...        | ...             | ...     | ...   | ...
\end{verbatim}
\end{AIboxC}

\noindent This example illustrates two difficulties simultaneously:
the missing-header case that forces the LLM to infer column semantics
before it can construct the header, and the cluster of near-miss shell
names with plausible but wrong prices that surrounds both target rows.

\subsection{LLM Column Proposal Mechanism}
\label{app:tabmwp_columns}

Column augmentation is performed by \texttt{gpt-oss-120b} via
the Together API.
To keep the prompt tractable regardless of how many distractor rows were
added, the LLM is shown only the \emph{original} (unaugmented) table for
domain context; the number of rows to fill, $n$, is derived from the
\emph{full} augmented table.

For each proposed column, the LLM supplies a short Python snippet that
builds a list called \texttt{values} with exactly $n$ elements.
The snippet may use \texttt{random}, \texttt{math}, and \texttt{string}
from the standard library, enabling varied but deterministic outputs
(seeded per entry).
A hard deduplication pass discards any proposed column whose name
(case-insensitive) already exists in the table, preventing the LLM from
reinstating existing columns under a slightly different spelling.
At most four new columns are accepted per entry.

For \textsc{Price List} tables the prompt is extended to ask for names
for the existing columns as well, using the format:

\begin{quote}
\small
\texttt{\{"existing\_column\_names": ["Shell Type", "Unit Price"],}\\
\texttt{\hspace{1em}"columns": [\{"column\_name": "Inventory Count",}\\
\texttt{\hspace{3em}"python\_code": "values = [random.randint(1,500) for \_ in range(n\_rows)]"\},}\\
\texttt{\hspace{1em}...]\}}
\end{quote}

\noindent The injected header is then built by concatenating the
existing-column names and the new-column names in order, ensuring a
well-formed header even for tables that were originally headerless.

\subsection{Collision Avoidance}
\label{app:tabmwp_collision}

Two independent mechanisms ensure that no distractor or near-miss row
introduces an ambiguous or incorrect ground truth.

\paragraph{Key-level collision guard.}
The distractor generators maintain the set of first-column keys present
in the original table and filter every candidate key against it before
inclusion.
For \textsc{Price List} tables, compound item names (e.g.\ ``apple bread''
from pairing ``apple'' and ``bread'' from the pool) prevent accidental
single-word matches.
For \textsc{Financial Ledger} tables, new transactions reuse only dates
already present in the original ledger; extending the date range would
shift ``end-of-period'' balance answers.
After all rows are merged, a deterministic validator scans the augmented
table and flags any original key that appears more than once; entries
with any such collision are discarded.

\paragraph{Answer-value near-miss validator.}
Near-miss rows are designed to confuse the model's \emph{row selection},
not to carry a numerically correct value.
After near-miss insertion, the LLM is given the list of newly added rows,
the question, and the ground-truth answer, and asked to issue a
\textsc{keep} or \textsc{remove} decision for each row according to
four criteria:
(i) the row's values would accidentally reproduce the correct answer
when used with the original table;
(ii) the row's first-column key already exists in the original table
(duplicate key);
(iii) the row's value format is inconsistent with the table (e.g.\
missing \$ signs in a currency table);
(iv) the row's label comes from a different semantic domain than the
other row labels (e.g.\ a sport name in a table of person names).
A rule-based pre-filter handles criterion (ii) before the LLM call,
removing exact-key duplicates without spending API tokens.
Rows flagged \textsc{remove} are deleted from the augmented table;
entries where no near-miss rows survive the filter are discarded
entirely rather than retained without adversarial pressure.

%% file: appendix/compute_cost.tex
\section{Compute and Training Cost}
\label{app:compute}

All SMITH training runs were performed on a single NVIDIA RTX 6000 Pro GPU. Each training session, covering one model size and one tool-pool configuration, took between 36 and 72 hours to complete depending on rollout length, number of pooled tools, and the size of the policy backbone (Qwen3-4B, Qwen3-8B, or Granite-3.3-8B). We report this to make the cost of reproducing our results explicit: the recipe is reachable on a single workstation-class GPU and does not require multi-node clusters.

%% file: appendix/trove.tex

\section{TroVE Baseline: A Controlled Prompt-Tuning Ablation}
\label{app:trove_tweak}

Appendix~\ref{app:trove_loop} describes TroVE's~\citep{wang2024trove} online
import/create/skip loop, and Appendix~\ref{app:baseline_failure} documents
implementation-level failure modes we found in TroVE and KTCE. A separate
concern is whether TroVE's prompt is simply \emph{under-tuned}: every
baseline in this paper uses that framework's own default, unmodified prompt,
so a reviewer could reasonably ask whether a small amount of prompt
engineering would close part of the gap to SMITH. We test this directly with
a controlled ablation on the shared \texttt{reasoning} prompt template (used
by every procedurally-generated reasoning task with no task-specific
overlay), run with Qwen3-4B-Instruct-2507 under TroVE's own frozen-toolbox
import/skip evaluation protocol (Appendix~\ref{app:trove_loop}).

\subsection{Versions compared}
\label{app:trove_tweak_versions}

\begin{itemize}[leftmargin=2em, itemsep=2pt]
  \item \textbf{v1 (baseline).} The original, unmodified prompt
        (\texttt{online\_create}/\texttt{online\_import}/\texttt{online\_skip}),
        byte-identical to what every non-ablation TroVE result elsewhere in
        this paper uses, and TroVE's registered default whenever no other
        version is explicitly requested.
  \item \textbf{v2 (bundled revision).} v1 plus three simultaneous changes:
        (a) an instruction to use plain ASCII punctuation, added after
        observing model-generated curly quotes triggering \texttt{SyntaxError}s
        at execution time; (b) an explicit ``if the toolbox is empty, still
        write code from scratch'' fallback sentence; and (c) a second worked
        example in \texttt{online\_import}.
  \item \textbf{v3 (punctuation only).} v1 plus \emph{only} change (a).
  \item \textbf{v4 (worked example only).} v1 plus \emph{only} change (c).
        Change (b) is a documented no-op under the current harness --- the
        injected toolbox is merged with a 3-entry stdlib-import seed set
        (\texttt{toolbox/reasoning.py}) before every call, so it is never
        actually empty --- and is omitted from both ablation arms.
\end{itemize}

\noindent Because v2 bundles three independent changes, an observed
regression cannot be attributed to any single one of them; v3 and v4 isolate
changes (a) and (c) respectively so the responsible change can be identified.

\subsection{Results}
\label{app:trove_tweak_results}

Table~\ref{tab:trove_prompt_ablation} reports strict exact-match accuracy for
all four versions on the eight reasoning tasks with complete, artifact-free
data (same frozen toolbox, same 100--350 test instances per task, paired by
question). Significance is a two-sided exact McNemar test on paired
correct/incorrect outcomes against v1.

\begin{table}[h]
  \centering
  \caption{TroVE prompt-tuning ablation (Qwen3-4B-Instruct-2507, frozen-toolbox
    import/skip evaluation, strict exact-match grading). $^{*}$: significantly
    different from v1 ($p<0.05$, two-sided exact McNemar test, paired by
    question). \texttt{caesar\_cipher}'s on-disk v2 run is an 8-example smoke
    test that predates the full sweep and is excluded as unreliable
    (``excl.''); its v1/v3/v4 columns all use the same full 196-question
    matched set. No configuration ever significantly \emph{outperforms} v1.}
  \label{tab:trove_prompt_ablation}
  \small
  \setlength{\tabcolsep}{5pt}
  \begin{tabular}{lrrrrr}
    \toprule
    \textbf{Task} & \textbf{N} & \textbf{v1} & \textbf{v2 (bundled)} & \textbf{v3 (punct.)} & \textbf{v4 (example)} \\
    \midrule
    bitwise\_arithmetic$^{a}$        & 280 &  55.0 & 27.5$^{*}$ & 40.4$^{*}$ & 45.4$^{*}$ \\
    caesar\_cipher                   & 196 &  59.2 & excl.      & 49.5$^{*}$ & 51.5$^{*}$ \\
    chinese\_theorem                 & 100 &  96.0 & 100.0      & 96.0       & 96.0       \\
    count\_bits                      & 210 & 100.0 & 100.0      & 100.0      & 100.0      \\
    isomorphic\_string               & 350 & 100.0 & 100.0      & 100.0      & 100.0      \\
    knights\_knaves                  & 210 &  51.0 & 50.5       & 51.0       & 52.9       \\
    polynomial\_equations            & 280 &  37.9 & 36.1       & 31.4$^{*}$ & 38.9       \\
    polynomial\_multiplication$^{a}$ & 275 &  52.7 & 36.4$^{*}$ & 45.8$^{*}$ & 51.3       \\
    \bottomrule
  \end{tabular}
\end{table}

Three of the eight tasks are already at or near ceiling for every version
(\texttt{count\_bits} and \texttt{isomorphic\_string} at 100.0\% throughout;
\texttt{chinese\_theorem} at 96--100\%, where v2's apparent 96.0\%$\to$100.0\%
jump is not significant, $p=0.125$, since it comes from only 4 errors total)
and \texttt{knights\_knaves} is flat across all four versions (50.5--52.9\%;
the largest deviation from v1 is v4's $+1.9$ points, $p=0.62$). On the
remaining four tasks, every \emph{significant} difference from v1 is a
regression: v3 and v4 each independently reproduce part of v2's damage on
\texttt{caesar\_cipher} (59.2\%$\to$49.5\%, $p=0.008$, and
$\to$51.5\%, $p=0.028$, respectively, with no significant difference between
v3 and v4 themselves, $p=0.67$) and on \texttt{polynomial\_multiplication}
(52.7\%$\to$45.8\%, $p=0.037$, for v3; v4's 51.3\% is not significantly
different from v1, $p=0.74$). On \texttt{polynomial\_equations}, only v3
regresses significantly (37.9\%$\to$31.4\%, $p=0.039$); v4 is statistically
indistinguishable from v1 ($p=0.78$). Whenever v3 and v4 differ, v4 (the
worked example alone) sits closer to v1 than v3 (the punctuation instruction
alone) or the full v2 bundle, suggesting the ASCII-punctuation instruction is
the larger contributor to v2's regression on the two polynomial tasks, while
both changes contribute comparably on \texttt{caesar\_cipher}.

\paragraph{Grading caveats.}
The two tasks marked $^{a}$ use exact-string-match grading that is sensitive
to formatting choices unrelated to reasoning quality. For
\texttt{polynomial\_multiplication}, re-grading with symbolic
(\texttt{sympy}) equivalence instead of exact string match raises every
version's accuracy (v1: 68.0\%, v2: 53.1\%, v3: 54.5\%, v4: 59.6\%) but
preserves the significance of the v1--v2, v1--v3, and v1--v4 gaps
($p=5.7\times10^{-6}$, $3.8\times10^{-5}$, and $0.028$). For
\texttt{bitwise\_arithmetic}, we found the entire strict-accuracy spread is
an artifact: predictions are marked wrong whenever they omit the gold
answer's \texttt{0x} hex prefix (e.g.\ gold \texttt{0x7975b8c1} vs.\ a
numerically-identical prediction \texttt{7975b8c1}). Re-grading by numeric
value instead of exact string, all four versions are statistically
indistinguishable and near-ceiling (v1: 99.3\%, v2: 99.3\%, v3: 97.5\%, v4:
100.0\%; $p\geq0.18$ for every version against v1), so
\texttt{bitwise\_arithmetic} in fact shows no genuine prompt effect at all.

\paragraph{Excluded tasks and a larger-scale spot check.}
Two further reasoning tasks, \texttt{tower\_of\_hanoi} and
\texttt{cryptarithm}, are omitted from Table~\ref{tab:trove_prompt_ablation}
because their evaluation logs repeat identical question text across many
underlying instances (e.g.\ the same ``3-disk Tower of Hanoi'' prompt appears
70 times), which collapses a question-keyed paired significance test down to
only 3 and 10 effectively distinct comparisons out of 210 logged instances
each --- too few to support any conclusion. As a higher-power spot check on
our largest reasoning task, GSM8K ($N=1319$, roughly $4$--$13\times$ larger
than any task in Table~\ref{tab:trove_prompt_ablation}), v1 reaches 91.1\%,
matched closely by v2's 90.2\% ($p=0.25$) and v3's 90.8\% ($p=0.78$) ---
consistent with the pattern above, no version we tested ever significantly
outperforms the original prompt.

\paragraph{Takeaway.}
Across every task where the comparison is statistically meaningful, neither
isolated change in v2 (nor v2 itself) ever significantly outperforms v1, and
on the tasks where v2 was known to regress, both isolated changes still
regress relative to v1, just less severely. This is why every non-ablation
TroVE result reported in this paper (Table~\ref{tab:rg_results},
Table~\ref{tab:ood_transposed}) uses the original v1 prompt: among the four
versions we tested, it is the strongest one available, so the gap between
TroVE and SMITH is not an artifact of an under-tuned TroVE prompt.